\documentclass[runningheads]{llncs}
\usepackage{graphicx}
\usepackage{comment}
\usepackage{amsmath,amssymb} 
\usepackage{color}

\usepackage[parfill]{parskip}
\begin{document}
\pagestyle{headings}
\mainmatter
\def\CVPPPSubNumber{22}  
\def\ECCVSubNumber{\CVPPPSubNumber} 
\title{Unsupervised Domain Adaptation For Plant Organ Counting} 

\titlerunning{Domain Adaptation for Organ Counting}
%
\author{Tewodros W. Ayalew \and
Jordan R. Ubbens \and
Ian Stavness}
\authorrunning{T. Ayalew et al.}
\institute{University of Saskatchewan, Saskatoon SK, S7N 5A8, Canada
\email{\{tewodros.ayalew,jordan.ubbens,ian.stavness\}@usask.ca}}
\maketitle

\begin{abstract}
   Supervised learning is often used to count objects in images, but for counting small, densely located objects, the required image annotations are burdensome to collect. Counting plant organs for image-based plant phenotyping falls within this category. Object counting in plant images is further challenged by having plant image datasets with significant domain shift due to different experimental conditions, e.g. applying an annotated dataset of indoor plant images for use on outdoor images, or on a different plant species. In this paper, we propose a domain-adversarial learning approach for domain adaptation of density map estimation for the purposes of object counting. The approach does not assume perfectly aligned distributions between the source and target datasets, which makes it more broadly applicable within general object counting and plant organ counting tasks. Evaluation on two diverse object counting tasks (wheat spikelets, leaves) demonstrates consistent performance on the target datasets across different classes of domain shift: from indoor-to-outdoor images and from species-to-species adaptation.
\end{abstract}

\section{Introduction}

Object counting is an important task in computer vision with a wide range of applications, including counting the number of people in a crowd~\cite{lempitsky2010learning,wang2015deep,ranjan2018iterative}, the number of cars on a street \cite{tayara2017vehicle,zhang2017fcn}, and the number of cells in a microscopy image~\cite{xie2018microscopy,zhu2017extended,paul2017count}. Object counting is a relevant task in image-based plant phenotyping, notably for counting plants in the field to estimate the rate of seedling emergence, and counting plant organs to estimate traits relevant for selection in crop breeding programs. For example, counting spikes or heads in cereal crops is a relevant trait for estimating grain yield~\cite{pound2017deep,lu2017tasselnet,alkhudaydi2019spikeletfcn}, counting flowers for estimating the start and duration of flowering is relevant for demarcating plant growth stages \cite{lin2018detection,aslahishahri2019kl,zhang2020image}, and counting leaves and tillers is a relevant trait to assess plant health \cite{giuffrida2018pheno,aich2017leaf}. Leaf counting, in particular, has been a seminal plant phenotyping task, thanks to the CVPPP leaf counting competition dataset \cite{MinerviniPRL2015,PlantPhenotypingDatasets2015}.

Convolutional neural networks (CNN) provide state-of-the-art performance for object counting tasks. Supervised learning is most common in previous work, but object detection \cite{li2008estimating,leibe2005pedestrian} and density estimation \cite{lempitsky2010learning,olmschenk2018crowd,lu2017tasselnet} approaches both require fairly tedious annotation of training images with either bounding boxes or dots centered on each object instance. In the context of plant phenotyping, plant organ objects are often small and densely packed making the annotation process even more laborious. In addition, unlike general objects which can be annotated reliably by any individual, identifying and annotating plant organs in images often requires specialized training and experience in plant science \cite{giuffrida2018citizen,zhou2018crowdsourcing}. This makes it difficult to obtain large annotated plant image datasets. 

Another challenge for computer vision tasks in plant phenotyping is that, unlike large image datasets of general objects, plant image datasets usually include highly self-similar images with a small amount of variation among images within the dataset. An individual plant image dataset is often acquired under similar conditions (single crop type, same field) and therefore trying to directly use a CNN trained on a single dataset to new images from a different crop, field, or growing season will likely fail. This is because a model trained to count objects on one dataset (source dataset) will not perform well on a different dataset (target dataset) when these datasets have different prior distributions. This challenge is generally called \emph{domain-shift}. An extreme case of domain shift in plant phenotyping is using a source dataset of plant images collected in an indoor controlled environment, which can be more easily annotated because plants are separated with controlled lighting on a blank background, and attempting to apply the model to a target dataset of outdoor field images with multiple, overlapping plants with variable lighting and backgrounds, and blur due to wind motion. Domain shift is often handled by fine-tuning a model, initially trained on a source dataset, with samples taken from a target dataset. Fine-tuning, however, still requires some annotated images in the target dataset. Another approach used to solve this problem is domain adaptation. Domain adaptation techniques typically try to align the source and the target data distributions \cite{valerio2019leaf}.

In this paper, we propose a method that applies an unsupervised domain adaptation technique, first proposed for image classification~\cite{ganin2015unsupervised}, to jointly train a CNN to count objects from images with dot annotations (i.e., source domain) and adapt this knowledge to related sets of images (i.e., target domain) where labels are absent. We modeled the object counting problem as a density map estimation problem.
We evaluated our proposed domain adaptation method on two object counting tasks (wheat spikelet counting and rosette leaf counting) each with a different challenge. The wheat spikelet counting task adapts from indoor images to outdoor images and presents challenges due to self-similarity, self-occlusion, and appearance variation \cite{pound2017deep,alkhudaydi2019spikeletfcn}. The leaf counting task adapts from one plant species to a different plant species and presents variability in leaf shape and size, as well as overlapping and occluded leaves \cite{kuznichov2019data}. Results show consistent improvements over the baseline models and comparable results to previous domain adaptation work in leaf counting. 
The contributions of our paper include: 1) the extension of an adversarial domain adaptation method for density map estimation that learns and adapts in parallel, 2) the evaluation of our method on two diverse counting tasks, and 3) a new public dataset of annotated wheat spikelets imaged in outdoor field conditions.

\section{Related Work}
\noindent\textbf{Object Counting}. A number of approaches have been proposed to train CNNs for object counting tasks. Among these, the most prominent methods are counting by regression~\cite{giuffrida2018pheno,dobrescu2017leveraging,ubbens2017deep,aich2017leaf}, counting by object detection~\cite{madec2019ear,gibbs2019recovering,ghosal2019weakly}, and counting by density estimation~\cite{lempitsky2010learning,gao2020cnn,sindagi2018survey,pound2017deep}. Here, we focus on density estimation methods, which estimate a count by generating a continuous density map and then summing over the pixel values of the density map. Density estimation uses weak labels (dot annotations on object centers), which are less tedious to annotate than bounding boxes or instance segmentation masks. Density estimation based counting was first proposed by Lempitsky \& Zisserman~\cite{lempitsky2010learning} who demonstrated higher counting accuracy with smaller amounts of data. 

\noindent\textbf{Spikelet Counting}. Pound et al.~\cite{pound2017deep} presented a method for counting and localizing spikes and spikelets using stacked hourglass networks. They trained and evaluated their model on spring wheat plant images taken in a controlled environment. Their approach achieved a $95.91\%$ accuracy for spike counting and a $99.66\%$ accuracy for spikelet counting. Even though their results show high accuracy, counting spikelets from real field images requires further fine-tuning of their model. Alkhudaydi et al.~\cite{alkhudaydi2019spikeletfcn} proposed a method to count spikelets from infield images using a fully convolutional network named SpikeletFCN. They compared the performance of training SpikeletFCN from scratch on infield images and fine-tuning the model, which was initially trained using images taken in a controlled setting~\cite{pound2017deep}. They also showed that manually segmenting the background increased performance, but manually annotating spikelets and manually removing backgrounds for field images is a time-consuming process.

\noindent\textbf{Leaf Counting}. Leaf counting in rosette plants has been performed with CNN-based regression approaches. Dobrescu et al.~\cite{dobrescu2017leveraging} used a pre-trained ResNet architecture fine-tuned on rosette images to demonstrate state-of-the-art leaf counting performance in the CVPPP 2017 leaf counting competition. Aich \& Stavness~\cite{aich2017leaf} presented a two-step leaf counting method, first using an encoder-decoder network to segment the plant region from the background, and then using a VGG-like CNN to estimate the count. Giuffrida et al.~\cite{giuffrida2018pheno} extended the CNN regression method to show that using multiple image modalities can improve counting accuracy. They also demonstrated the generality of the approach by evaluating with different plant species and image modalities. Itzhaky et al.~\cite{itzhaky2018leaf} proposed two deep learning based approaches that utilize a Feature Pyramid Network (FPN) for counting leaves, namely using direct regression and density map estimation. Their density map based approach achieved $95\%$ average precision.


\noindent\textbf{Domain Adaptation}. In recent years, a number of deep domain adaptation approaches have been proposed to minimize the domain shift between a source dataset and a target dataset~\cite{hu2015deep,tzeng2015simultaneous,long2017deep,bousmalis2017unsupervised,tzeng2017adversarial,liu2016coupled}. Tzeng et al.~\cite{tzeng2017adversarial} proposed an adversarial approach for an unsupervised domain adaptation named Adversarial Discriminative Domain Adaptation (ADDA). ADDA minimizes the representation shift between the two domains using a Generative Adversarial Network (GAN) loss function where the domains are allowed to have independent mappings. As an alternative to a GAN loss, Ganin \& Lempitsky~\cite{ganin2015unsupervised} presented an unsupervised domain adaptation technique by creating a network that has a shared feature extraction layers and two classifiers. The adaptation process works by minimizing the class label prediction loss and maximizing the domain confusion. These previous works have only been evaluated on image classification problems. Domain adaptation has garnered less attention for other computer vision tasks, such as density estimation or regression problems.

\noindent\textbf{Domain Adaptation in Plant Phenotyping}. Giuffrida et al.~\cite{valerio2019leaf} proposed a method to adapt a model trained to count leaves in a particular dataset to an unseen dataset in an unsupervised manner. Under the assumption that the images in the source domain are private, their method uses the representations of images from the source domain (from a pretrained network) to adapt to the target domain. They employed the Adversarial Discriminative Domain Adaptation approach to solving the domain shift problem in the leaf counting task across different datasets. Their method aligns the predicted leaf count distribution with the source domain's prior distribution, which limits the adapted model to learning a leaf count distribution similar to the source domain. They have shown their method achieved a mean square error of $2.36$ and $1.84$ for intra-species and inter-species domain adaptation experiments, respectively. For fruit counting in orchard images, Bellocchio et al.~\cite{bellocchio2020combining} proposed a Cycle-GAN based domain adaptation method combined with weak presence/absence labels. They demonstrated state-of-the-art counting results for many combinations of adapting between plant species (almond, apple, olive). 

These promising results on leaf and fruit counting motivate a broader investigation of different domain adaptation approaches, which could be used for counting different plant organs and different categories of domain shift, e.g. from indoor image to field images. 
To the best of our knowledge, this is the first work that has applied domain adaptation to wheat spikelet counting and to a counting task with an indoor-to-outdoor domain shift.

\section{Method}
In this section, we discuss the problem setting for our proposed method, followed by a description of the proposed architecture and cost function. Finally we describe the training procedures used for all experiments.

\subsection{Problem Setting}
We model the task of domain adaptation in object counting as an unsupervised domain adaptation problem. We assume that the data is sampled from two domains: a source domain $(\mathcal{D}^s)$ and a target domain $(\mathcal{D}^t)$. We also assume that labels only exist for the images sampled from the source domain. Therefore, the source domain is composed of a set of images ($\mathcal{X}^s$) and their corresponding labels ($\mathcal{Y}^s$). Whereas, the target domain only has images ($\mathcal{X}^t$) without labels.

\subsection{Architecture}
\begin{figure*}
\begin{center}
\includegraphics[width=\textwidth]{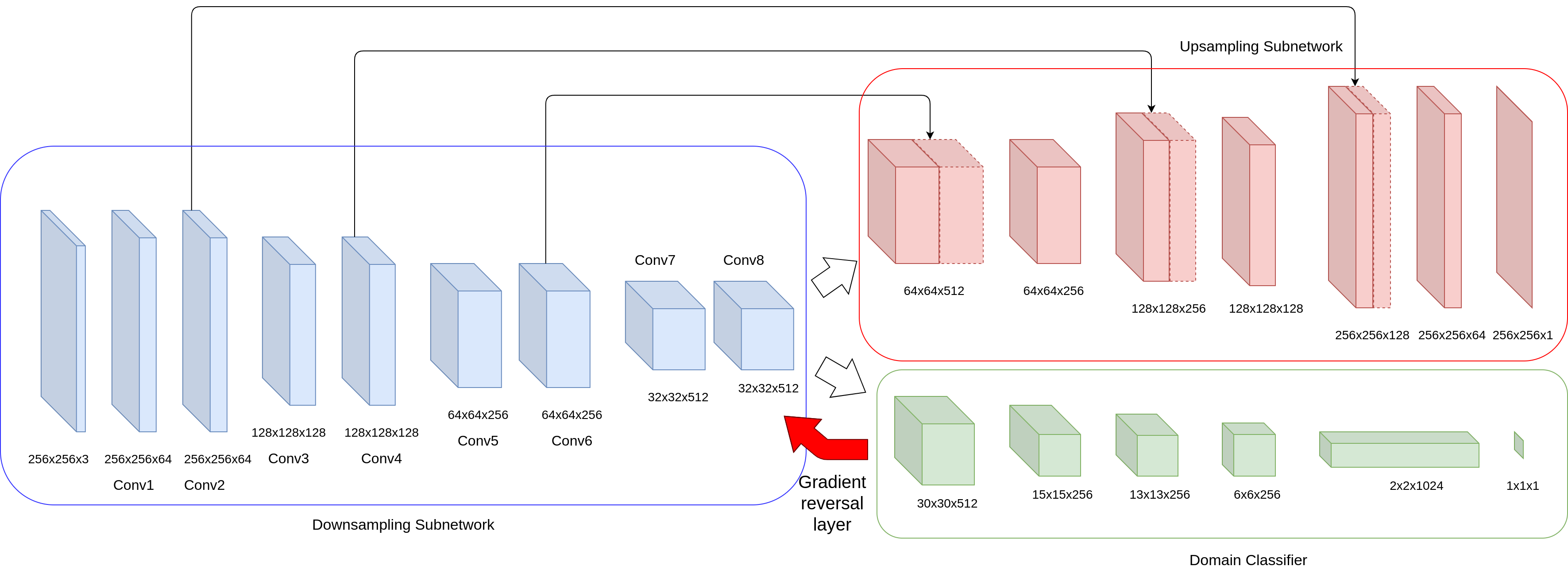}
\end{center}
   \caption{The proposed Domain-Adversarial Neural Network composed of two networks that share weights between Conv1 and Conv8. The downsampling subnetwork~$(G_d)$, the upsampling subnetwork~$(G_u)$, and the domain classifier~$(G_c)$ are denoted by the blue, red, and green boxes respectively. The red arrow shows the Gradient Reversal Layer used to reverse the gradient in the backpropagation step as proposed by \cite{ganin2015unsupervised}.}
\label{fig:architecture}
\end{figure*}

Our proposed model is a Domain-Adversarial Neural Network (DANN), a class of models designed for unsupervised domain adaptation tasks \cite{ganin2016domain}. These networks are commonly characterized by two parallel classification networks that share weights in the first $n$ layers and a custom layer called Gradient Reversal Layer (GRL) \cite{wang2018deep}. One of the parallel networks is designed for the main classification task, while the second network is a classifier designed to discriminate whether an input is sampled from the source domain or from the target domain. We customize this general architecture by replacing the main classification network with a U-Net network \cite{ronneberger2015u} that is used for density map estimation.

Our proposed architecture is a fully convolutional network composed of two parallel networks, as can be seen in Figure~\ref{fig:architecture}. These parallel networks share weights from \emph{Conv1} to \emph{Conv8}. Given these shared weights, the architecture can be seen as one network containing three subnetworks: the downsampling subnetwork $(G_d)$, the upsampling subnetwork $(G_u)$, and the domain classifier subnetwork $(G_c)$. In Figure~\ref{fig:architecture} these subnetworks are denoted by different color boxes. The model takes an image $x_j$ and predicts a density map as $\hat{y}_{density} = G_u(G_d(x_j))$ and a class probability prediction representing whether $x_j$ is sampled from the source domain or from the target domain given by $\hat{y}_{domain} = G_c(G_d(x_j))$. 

A downsampling subnetwork followed by an upsampling subnetwork (sometimes referred to as an Hourglass network or encoder-decoder network) is a commonly used architecture for density estimation tasks\cite{liu2018crowd,xie2018microscopy}. The general structure for the Hourglass network we used is adapted from U-Net \cite{ronneberger2015u} with slight modifications. The downsampling subnetwork is composed of blocks of two $3\times3$ padded convolutions followed by a $2\times2$ max pooling layer with a stride of 2. The upsampling network is composed of blocks of $2\times2$ transpose convolutions followed by a pair of $3\times3$ padded convolutions. The feature maps generated from each convolution preceded by a transpose convolution are concatenated with their corresponding feature maps from the upsampling subnetwork (shown in Figure~\ref{fig:architecture} as dashed boxes). The domain classifier subnetwork is formed from $3\times3$ unpadded convolutions, a $2\times2$ maxpooling layers, and a $1\times1$ convolution (which is used to reduce the dimension of the output to a single value). All of the convolutions in our network, with the exception of the ones in the output layers, are followed by batch normalization and a ReLU activation. The output layer is followed by a sigmoid activation function.

The output of the downsampling subnetwork passes through a Gradient Reversal Layer (GRL) before passing through to the domain classifier subnetwork. The GRL, a custom layer proposed by \cite{ganin2015unsupervised}, is designed to reverse the gradient in the backpropagation step by multiplying the gradient by a negative constant ($-\lambda$). The purpose of reversing the gradient is to force the network into finding domain-invariant features common to both the source- and the target- domain. This aligns the distributions of the feature representations of images sampled from the two domains. 

\subsection{Cost Function}

The parameters in our proposed model are optimized by minimizing a two-part cost function, as follows:
\begin{align}
\mathcal{L}_{density} &= \log(\frac{1}{N}\sum_{i=0}^{N} (G_u(G_d(x^{s}_{i})) - y^{s}_{i})^2) \\[2pt]
\mathcal{L}_{domain} &= -\mathbb{E}_{x^s \sim \mathcal{X}^s}[\log(G_c(G_d(x^{s}_{i}))] 
         -\mathbb{E}_{x^t \sim \mathcal{X}^t}[1-\log(G_c(G_d(x^{t}_{i}))] \\[3pt]
\mathcal{L} &= \mathcal{L}_{density} + \mathcal{L}_{domain} 
\label{eqn:1}
\end{align}
where $(x^{s}_{i},y^{s}_{i})$ represents the $i$th image and density map label sampled from the source domain ($\mathcal{D}^s$).
The first part ($\mathcal{L}_{density}$) accounts for errors in the density map estimation for samples taken from the source domain. This is implemented as the logarithm of mean square error between the ground truth sampled from the source dataset and the predicted density map from the model. The second part ($\mathcal{L}_{domain}$) is a Binary Cross-Entropy loss for domain classification.

\subsection{Training}

Conceptually, the general training scheme we use to train our model can be decomposed into two sub-tasks: supervised density estimation on the source dataset, and domain classification on the merged source and target datasets. The goal of this training scheme is to extract relevant features for the density estimation task and, in parallel, condition the downsampling network to extract domain-invariant features.

In all experiments, we randomly partitioned the source domain dataset into training and validation sets. Since we only have ground truth density maps for the source domain, in-training validations are carried out using the validation set taken from the source domain. 

The model was trained with a batch size of 8. The input images in the training set were resized to $256\times 256$ pixels. We used the Adam optimizer with tuned learning rates for each subnetwork. We used a learning rate of $10^{-3}$ for the downsampling and upsampling subnetworks, and $10^{-4}$ for the domain classifier.

The constant parameter $\lambda$ --- which multiplies the gradient in the GRL layer --- is updated on each iteration during training. Following~\cite{ganin2016domain}, the value is provided to the network as a scheduler in an incremental manner ranging from 0 to 1. 

\section{Experiments}

In this section, we present two object counting tasks to evaluate the proposed method: wheat spikelet counting and leaf counting. 

Our proposed method's performance in all of the experiments is evaluated using target domain data as our test set. Our experiments are mainly assessed using Root Mean Square Error~(RMSE) and Mean Absolute Error~(MAE):  
\[ RMSE = \sqrt{\dfrac{1}{N}\sum\limits_{i=0}^{N}(\hat{y_i}-y_i)^2}
\]
\[
 MAE = \dfrac{1}{N}\sum\limits_{i=0}^{N}\left|\hat{y_i}-y_i\right|
\]
where $N$ represents the number of test images, $\hat{y_i}$ and $y_i$ represent the predicted value and the ground truth for the $i$th sample, respectively.

In addition to these metrics, we include task-specific metrics that are commonly used in the literature for each task. For the wheat spikelet task, we provide coefficient of determination ($R^2$) between the predicted and ground truth counts. For the leaf counting task, we report metrics that are provided for the CVPPP 2017 leaf counting competition. These metrics are Difference in Count ($DiC$), absolute Difference in Count ($|DiC|$), mean squared error ($MSE$) and percentage agreement($\%$).

To demonstrate the counting performance improvement provided by domain adaptation, we use a vanilla U-Net as a baseline model by removing the domain classifier sub-network from our proposed architecture. We train this baseline model exclusively on the source domain and evaluate performance on the target domain data. All experiments were performed on a GeForce RTX 2070 GPU with 8GB memory using the Pytorch framework. The implementation is available at: \url{https://github.com/p2irc/UDA4POC}

\begin{figure}[t]%
\begin{center}
\begin{tabular}{ccc}
  \includegraphics[width=.25\textwidth]{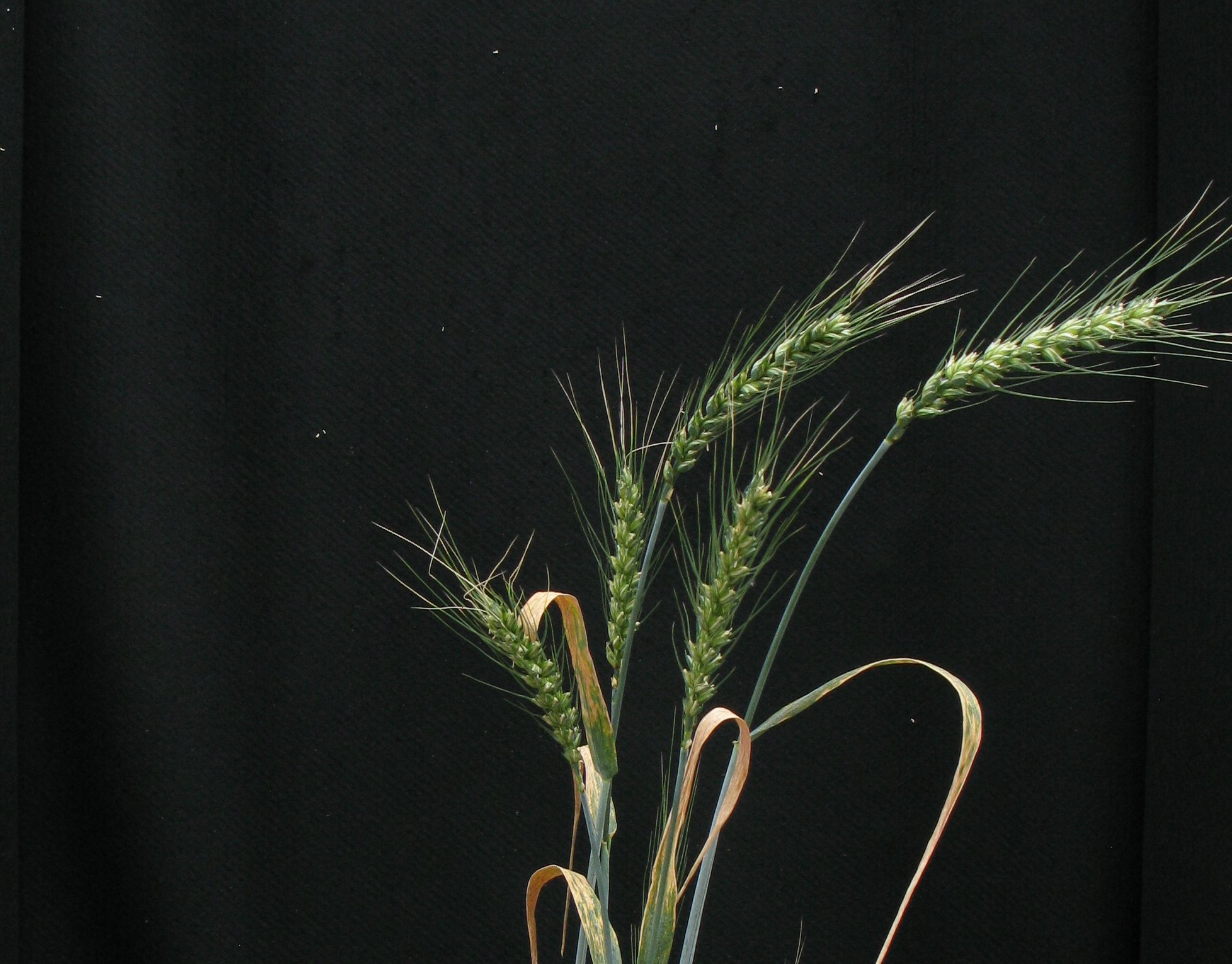} &
  \includegraphics[width=.25\textwidth]{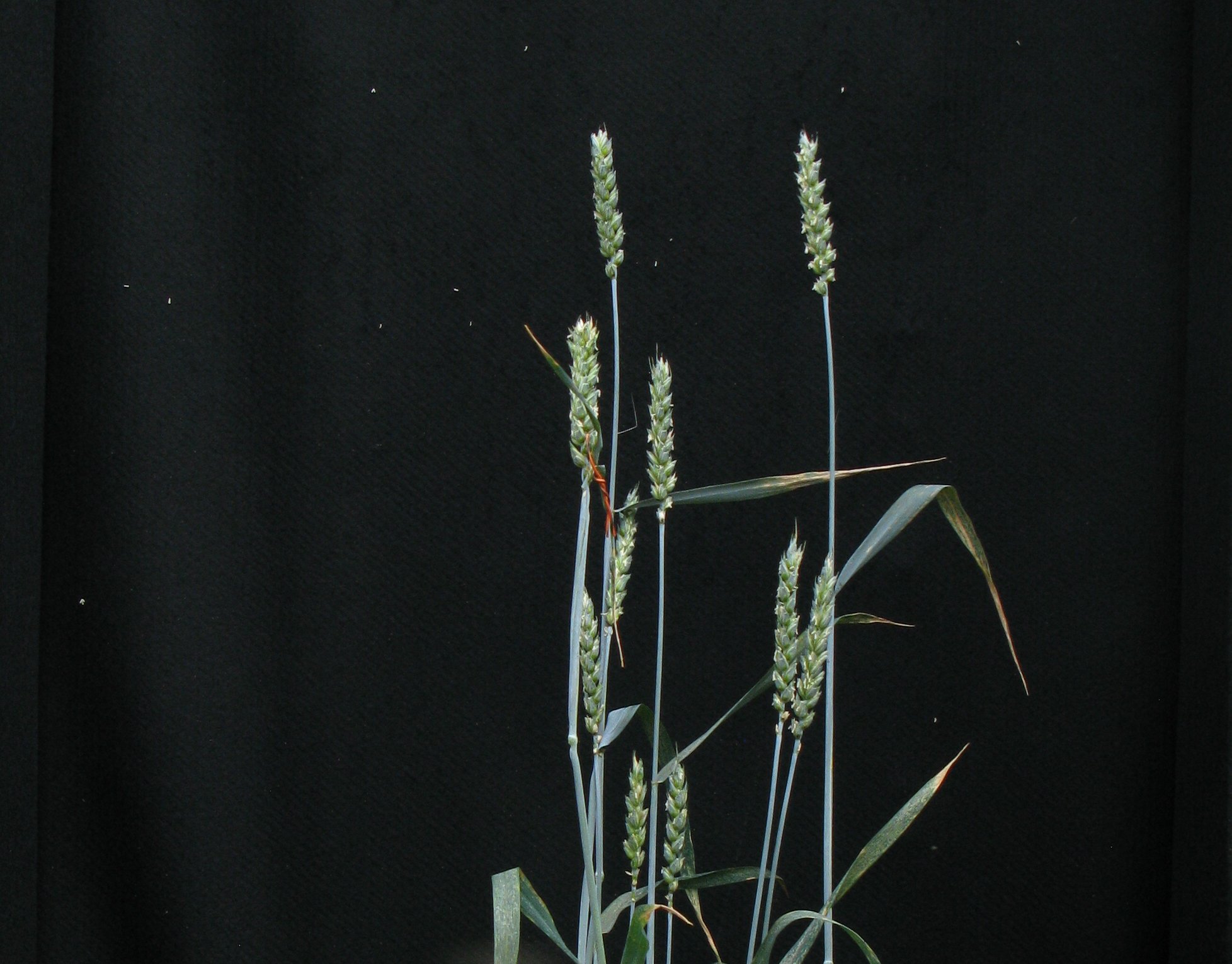} &
  \includegraphics[width=.25\textwidth]{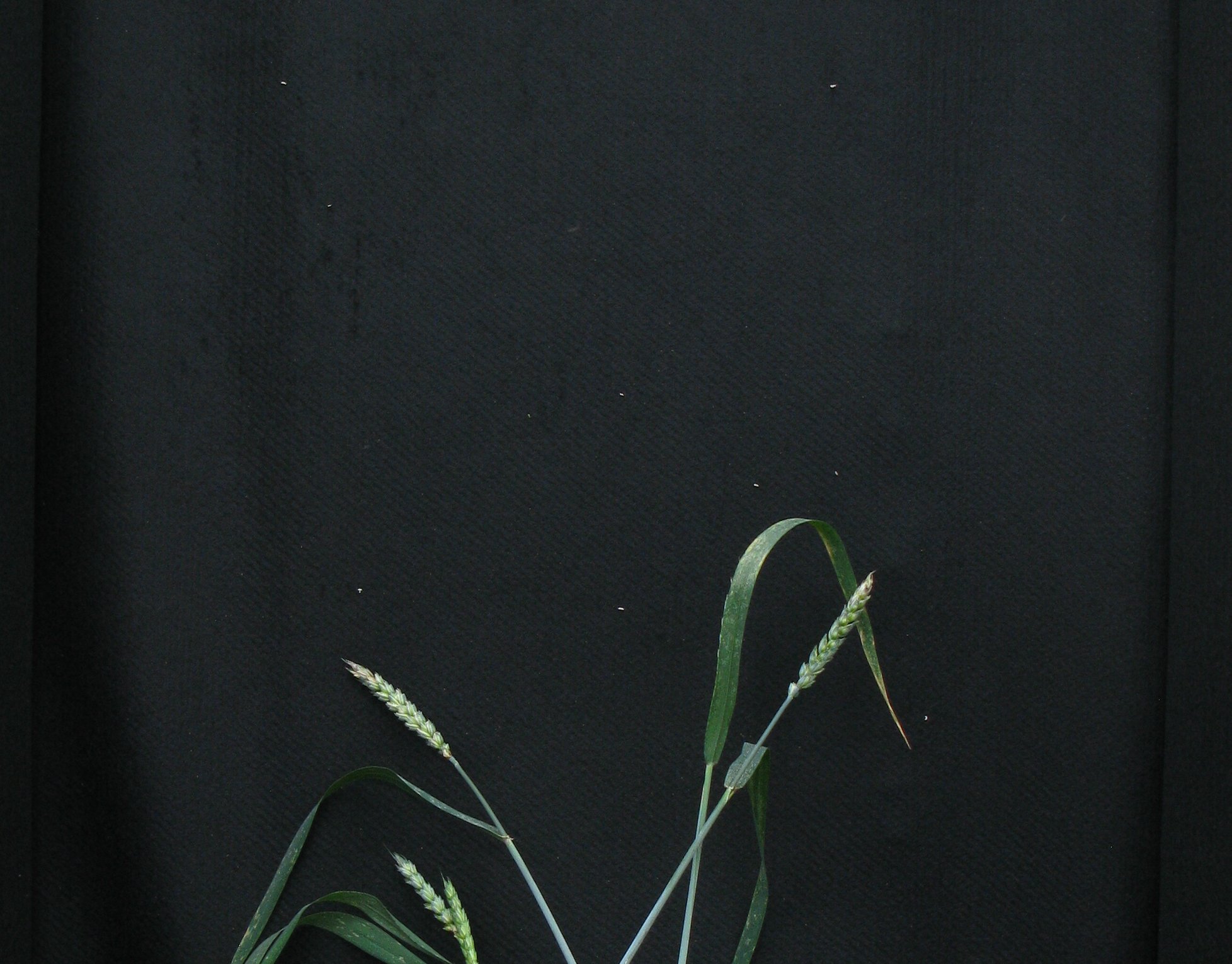} \\
  \includegraphics[width=.25\textwidth]{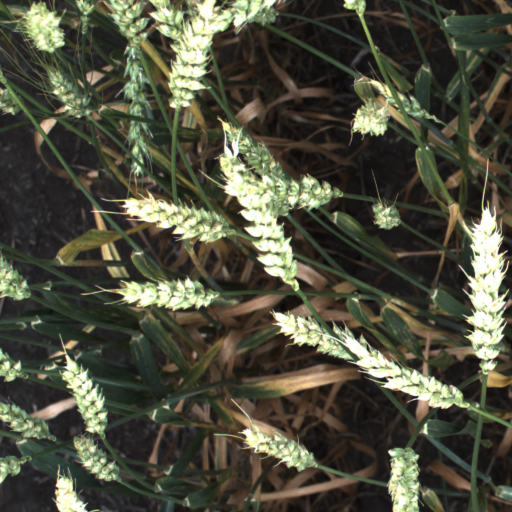} &
  \includegraphics[width=.25\textwidth]{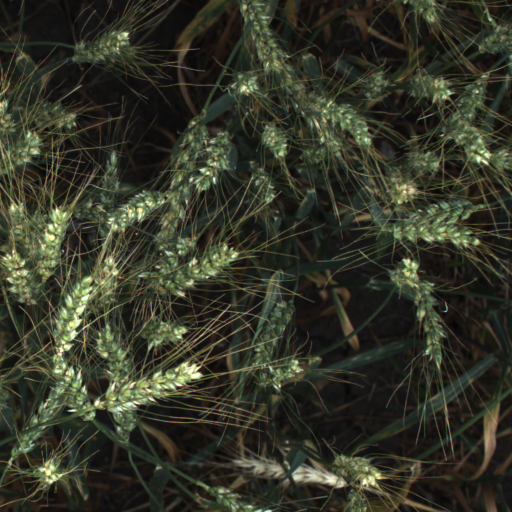} &
  \includegraphics[width=.25\textwidth]{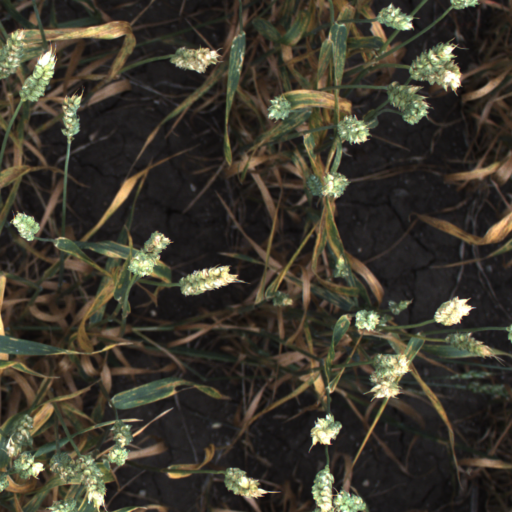}\\
  \includegraphics[width=.25\textwidth]{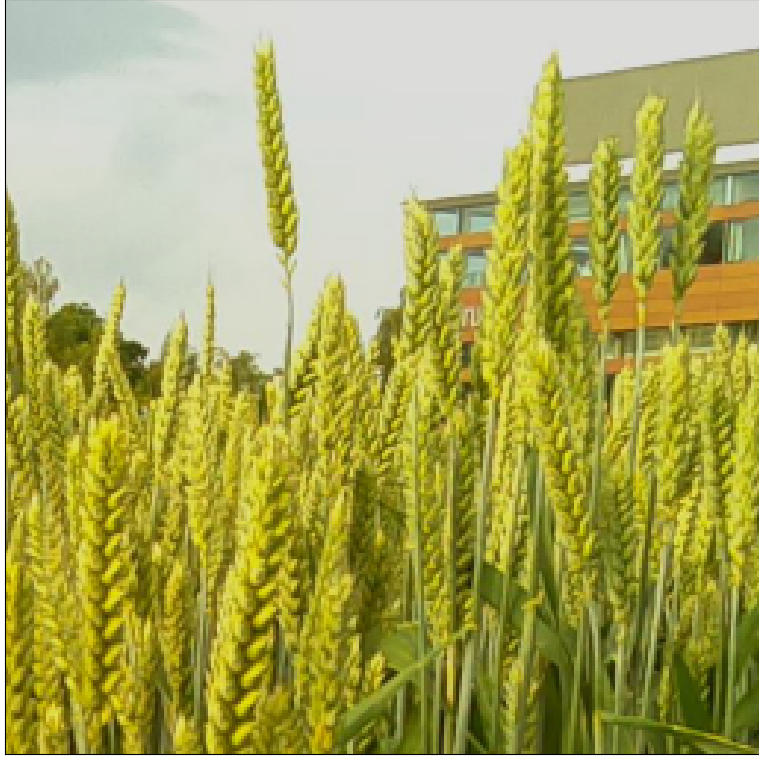} &
  \includegraphics[width=.25\textwidth]{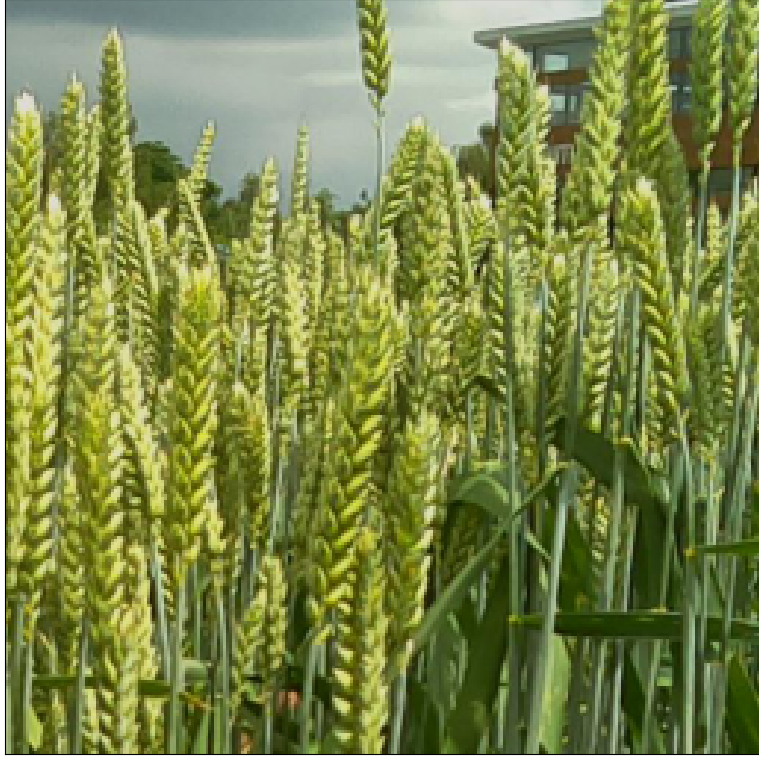} &
  \includegraphics[width=.25\textwidth]{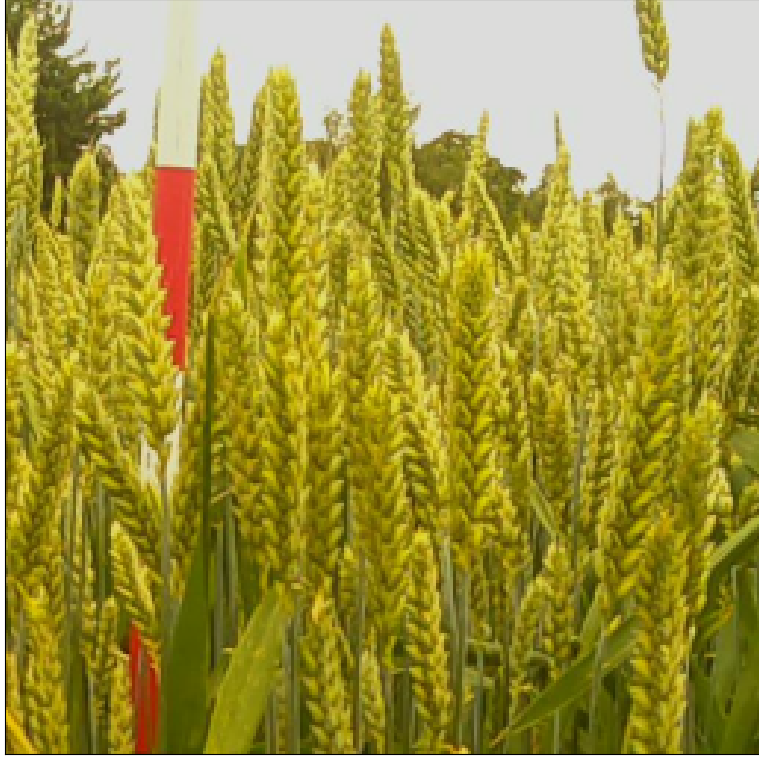}
\end{tabular}%
\end{center}
\caption{Example images for wheat spikelet counting experiments. Top: Source dataset, ACID \cite{pound2017deep}. Middle: Target dataset 1, Global Wheat Head Dataset \cite{david2020global}. Bottom: Target dataset 2, CropQuant dataset \cite{zhou2017cropquant}.}
\label{fig:wheat_source_target_sample}
\end{figure} 

\subsection{Wheat Spikelet Counting}

For the wheat spikelet counting task, we used the general problem definition of wheat spikelet counting as stated in~\cite{alkhudaydi2019spikeletfcn}.
Under this setting, our second experiment tries to adapt the wheat spikelet counting task from images taken in a greenhouse to images captured in an outdoor field. The datasets used for this task are as follows:

\begin{enumerate}
    \item \emph{ACID dataset} \cite{pound2017deep}: This dataset is composed of wheat plant images taken in a greenhouse setting. Each image has dot annotations representing the position of each spikelet. The ground truth density maps were generated by filtering the dot annotations with a 2D gaussian filter with $\sigma=1$.
    \item \emph{Global Wheat Head Dataset} (GWHD) \cite{david2020global}: A dataset presented in the Kaggle wheat head detection competition. The dataset comes with bounding box annotations for each wheat head, but these are not used in the present study.
    \item \emph{CropQuant dataset} \cite{zhou2017cropquant}: This dataset is composed of images collected by the Norwich Research Park (NRP). 
    As a ground truth, we used the dot annotations created by \cite{alkhudaydi2019spikeletfcn}, where they annotated 15 of the images from the CropQuant dataset. The ground truth has a total of 63,006 spikelets.
\end{enumerate}

We chose the ACID dataset as the source domain and the GWHD dataset as the target domain. Sample images from these datasets are displayed in Figure~\ref{fig:wheat_source_target_sample}. We randomly sampled 80\% of the images in the source dataset for training. These images were mixed with 200 images taken from the target dataset, making up the training set. Using this training set, the network is trained on the ACID dataset for spikelet counting and, in parallel, adapted to the GWHD. To evaluate our method, we created dot annotations for 67 images from the GWHD which are used as ground truth. These annotations are made publicly available at \url{https://doi.org/10.6084/m9.figshare.12652973.v2}.

In the second experiment, we used the CropQuant dataset as the target domain. To be consistent with the experiments presented in \cite{alkhudaydi2019spikeletfcn}, we randomly extracted $512\times512$ sized patches from each image in the CropQuant dataset. We generated 300 patches as the target domain. We partitioned the source domain with 80:20 training-validation split. We merged the training split with the generated patches and trained the model. 

\begin{figure}[ht]
\begin{center}
\begin{tabular}{ccc}
  \includegraphics[width=.25\textwidth]{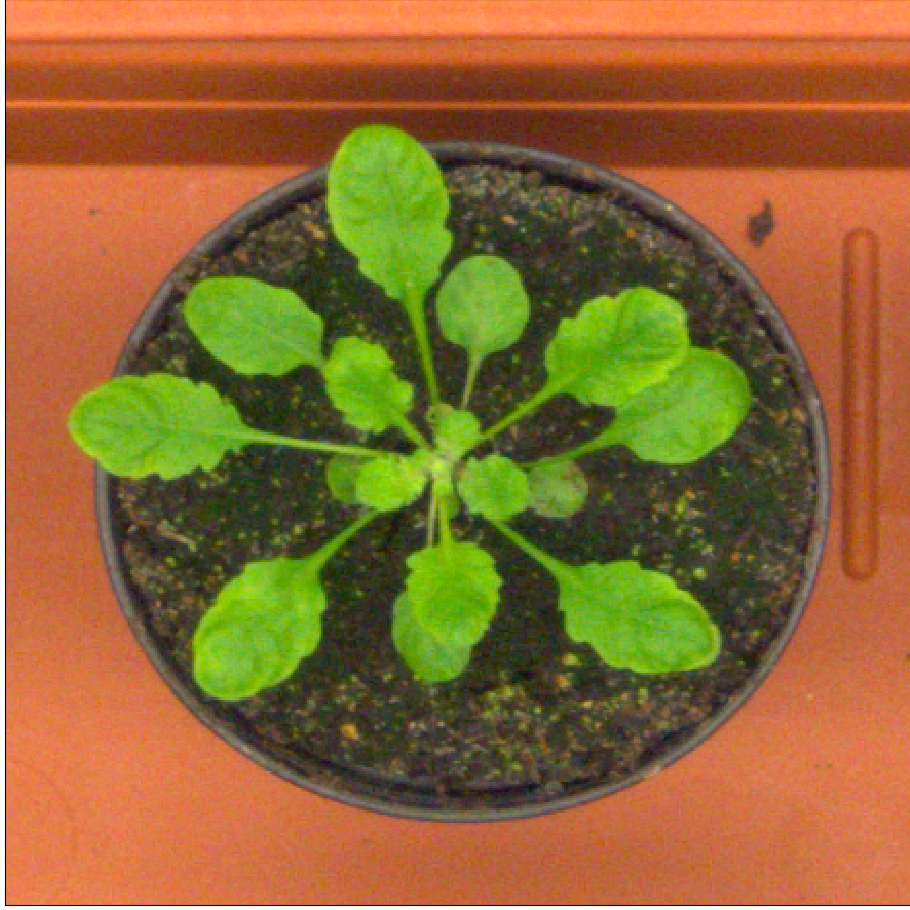} & 
  \includegraphics[width=.25\textwidth]{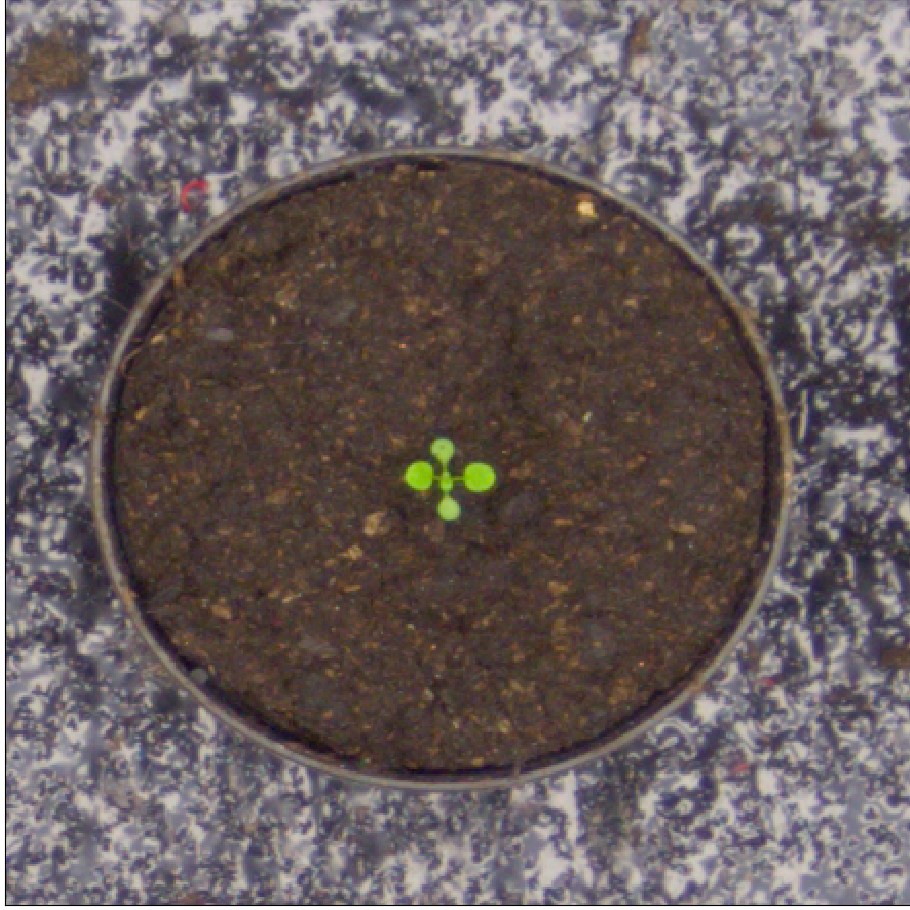} &
  \includegraphics[width=.25\textwidth]{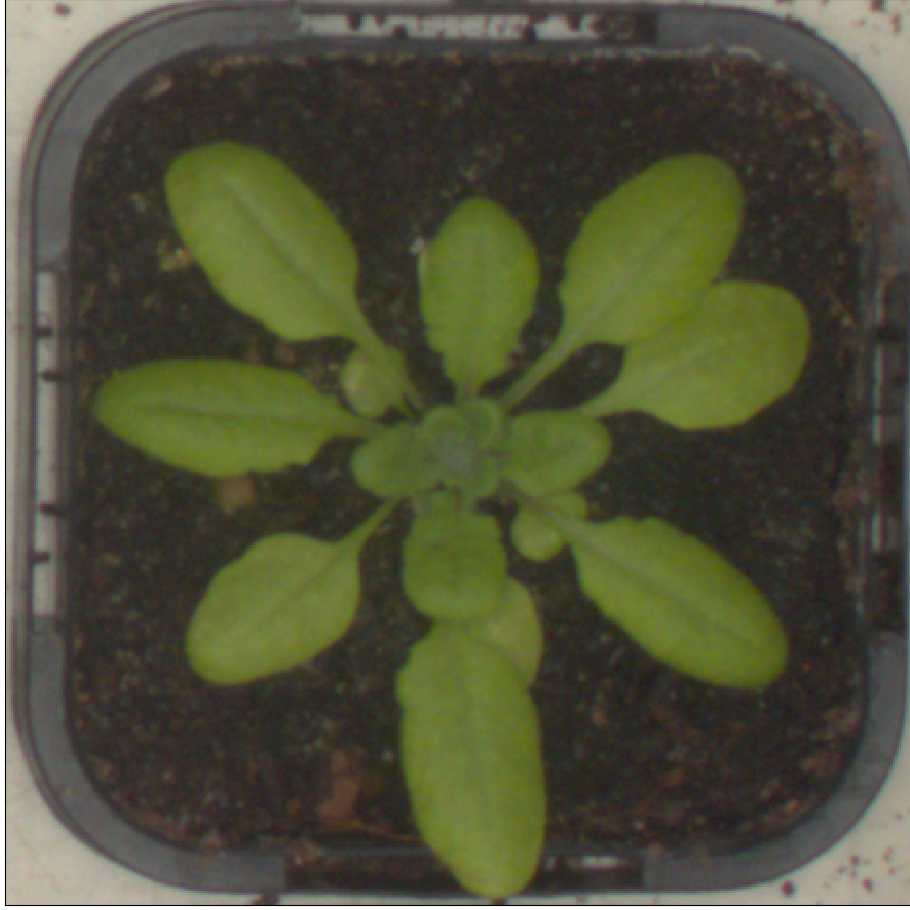} \\
  \includegraphics[width=.25\textwidth]{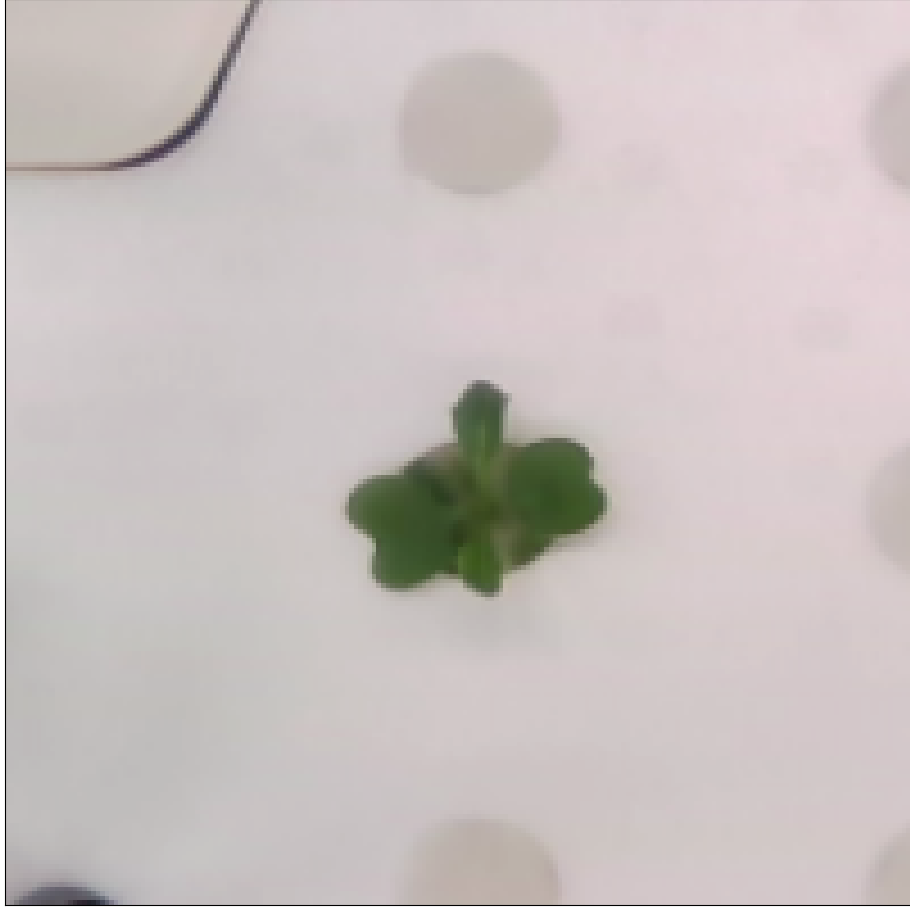} &
  \includegraphics[width=.25\textwidth]{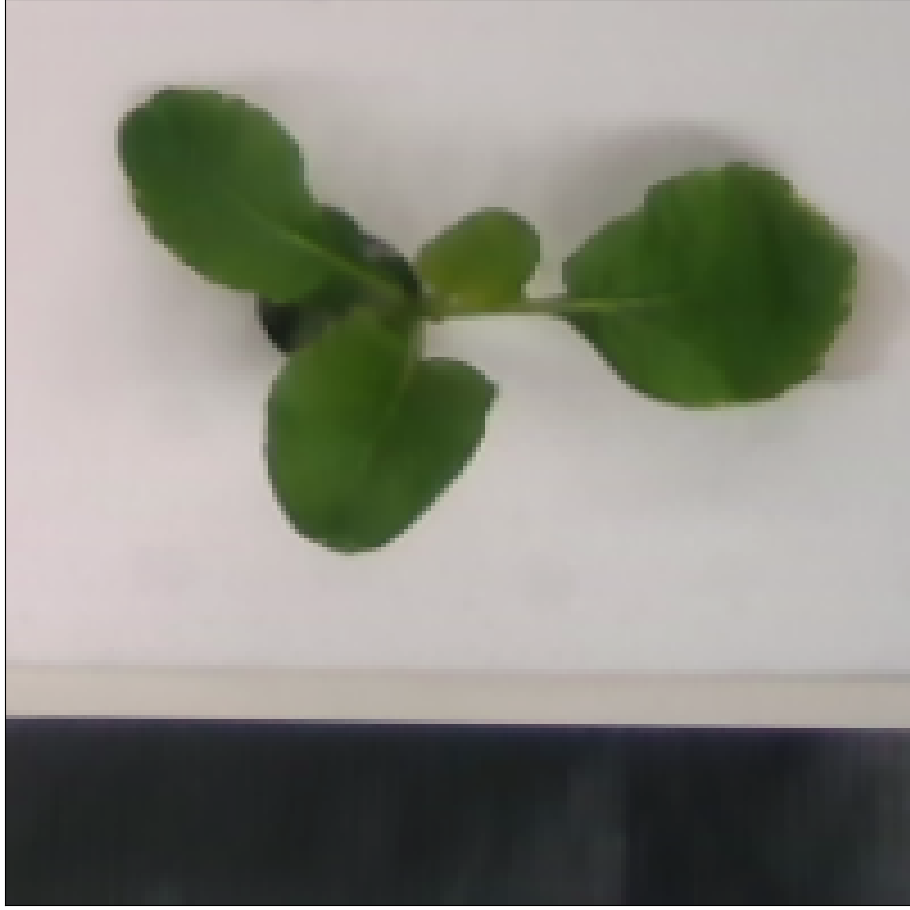} &
  \includegraphics[width=.25\textwidth]{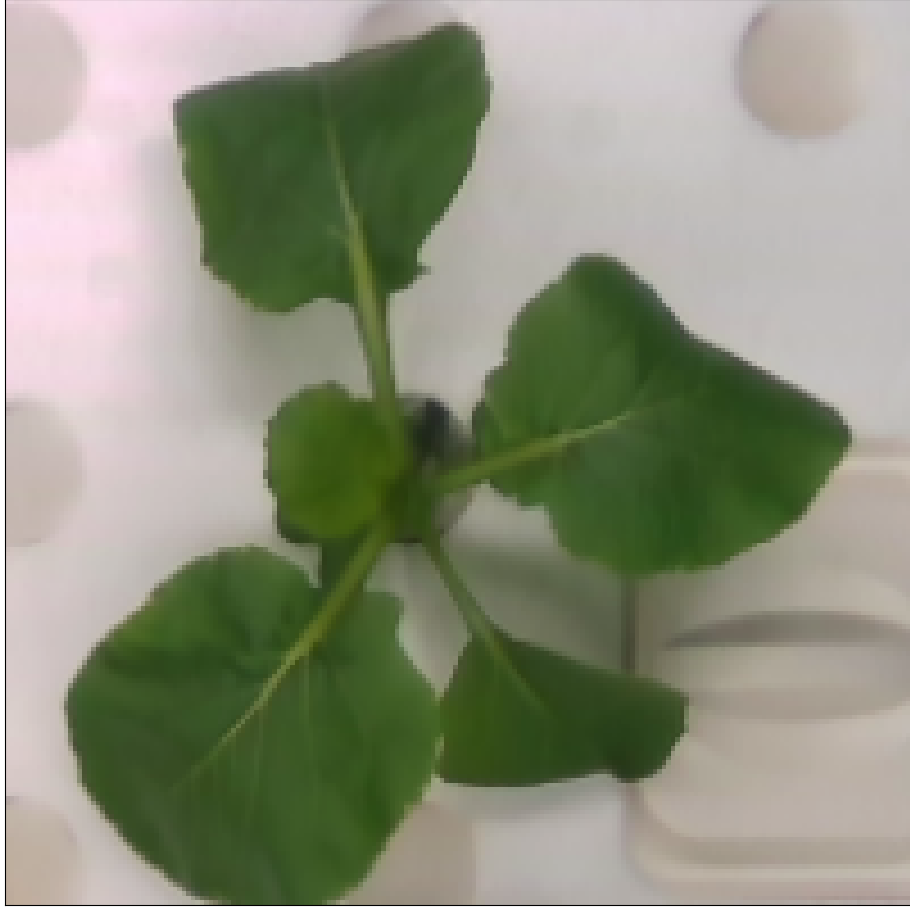} \\
  \includegraphics[width=.25\textwidth]{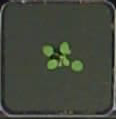} &
  \includegraphics[width=.25\textwidth]{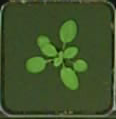} &
  \includegraphics[width=.25\textwidth]{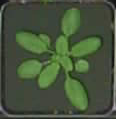}
 \end{tabular}
\end{center}

\caption{Example images for leaf counting experiments. Top: Source dataset, \emph{Arabidopsis}, CVPPP Dataset~\cite{MinerviniPRL2015,PlantPhenotypingDatasets2015}. Middle: Target dataset 1, KOMATSUNA Dataset~\cite{uchiyama2017easy}. Bottom: Target dataset 2, \emph{Arabidopsis} MSU-PID Dataset~\cite{cruz2016multi}.}
\label{fig:leaf_source_target_sample}
\end{figure}

\subsection{Leaf Counting}
For the leaf counting task, we follow~\cite{valerio2019leaf} in order to compare to their baseline results. Therefore, we used the following datasets:

\begin{enumerate}
    \item \emph{CVPPP 2017 LCC Dataset} \cite{MinerviniPRL2015,PlantPhenotypingDatasets2015}: A dataset compiled for the CVPPP leaf counting competition, containing top-view images of \emph{Arabidopsis thaliana} and tobacco plants. To be consistent with \cite{valerio2019leaf}, we use the A1, A2 and A4 \emph{Arabidopsis} image subsets. The dataset includes annotations for leaf segmentation masks, leaf bounding boxes, and leaf center dot annotations. We use the leaf center annotations and generate the groudtruth density map by filtering the center points with a 2D gaussian filter with $\sigma=3$.
    \item \emph{KOMATSUNA Dataset} \cite{uchiyama2017easy}: Top-view images of Komatsuna plants developed for 3D plant phenotyping tasks. We use the RGB images for the experiments presented here.
    \item \emph{MSU-PID}\cite{cruz2016multi}: A multi-modal imagery dataset which contains images of \emph{Arabidopsis} and bean plants. Among these, we used the RGB images of the \emph{Arabidopsis} subset.
\end{enumerate}

Figure~\ref{fig:leaf_source_target_sample} shows sample images from these datasets. Using these datasets, we performed three leaf-counting experiments in total. In all experiments, the CVPPP dataset was used as the source domain. 

The first experiment involved adapting leaf counting from the CVPPP dataset to the KOMATSUNA dataset. For this experiment, we randomly selected $80\%$ of the images from the CVPPP dataset and merged them with the images from the KOMATSUNA dataset. The remaining $20\%$ of the source domain is used for validation. With this setting, we trained our model for 150 epochs. Finally, we evaluated performance on the images from the KOMATSUNA dataset.

In the second experiment, we used the MSU-PID dataset as the target domain. The data from the source domain was partitioned into training and validation sets in a similar procedure as experiment 1. The model was evaluated using the MSU-PID dataset.

In the final experiment, we aimed to verify that our domain adaptation scheme is independent of the leaf count distributions in the source domain. To test this, we took the model trained in the first experiment and evaluated it on composite images created from KOMATSUNA containing multiple plants. The composite images are generated by randomly selecting four images from the KOMATSUNA dataset and stitching them together form a $2\times2$ grid. 

\section{Results}

In this section, we present the performance of the proposed model on two counting tasks. We also compare these results with the baseline model and with other existing methods, where possible.

\setlength{\tabcolsep}{4pt}
\begin{table}
\begin{center}
\caption{Domain adaptation results for the wheat spikelet counting tasks.}
\label{table:wheat_counting_results}
\begin{tabular}{lllc}
\hline\noalign{\smallskip}
& $MAE$ & $RMSE$ & $R^2$ \\
\noalign{\smallskip}
\hline
\noalign{\smallskip}
\textbf{ACID to GWHD}\\
\noalign{\smallskip}
Baseline (No Adaptation)  & 91.65 & 114.4 & 0.005 \\
Our method & 29.48 & 35.80 & 0.66\\
\hline
\noalign{\smallskip}
\textbf{ACID to CropQuant}\\
\noalign{\smallskip}
Baseline (No Adaptation)  & 443.61 & 547.98 & -1.70 \\
SpikeletFCN~\cite{alkhudaydi2019spikeletfcn} (Scratch) & 498.0 & 543.5 & -\\
SpikeletFCN~\cite{alkhudaydi2019spikeletfcn} (Fine-tuned) & 77.12 & 107.1 & -\\
Our method & 180.69 & 230.12 & 0.43\\
\noalign{\smallskip}
\hline
\end{tabular}
\end{center}
\end{table}
\setlength{\tabcolsep}{1.4pt}

\subsection{Wheat Spikelet Counting}
\begin{figure}[!ht]
\begin{center}
\begin{tabular}{cccc}
(a) Input & (b) GT & (c) Baseline & (d) Ours \\
  \includegraphics[width=.23\textwidth]{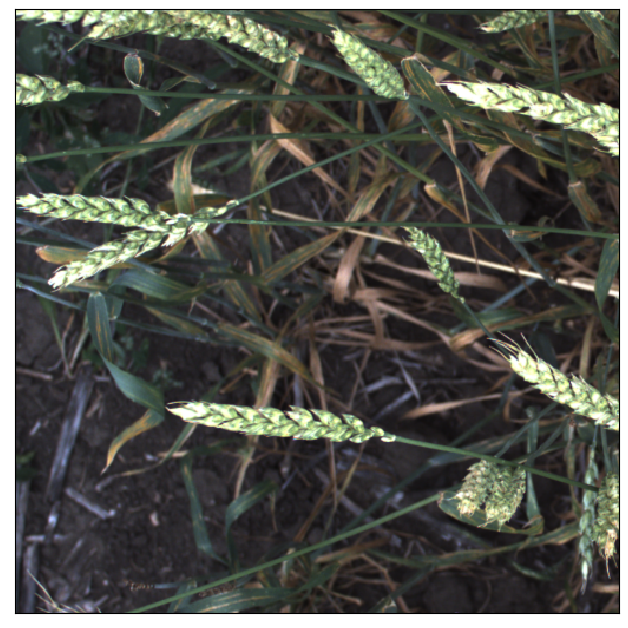} &
  \includegraphics[width=.23\textwidth]{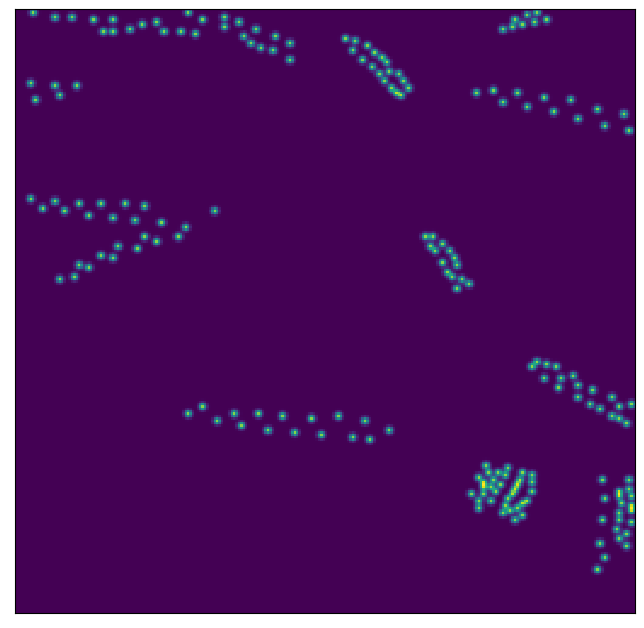} &
  \includegraphics[width=.23\textwidth]{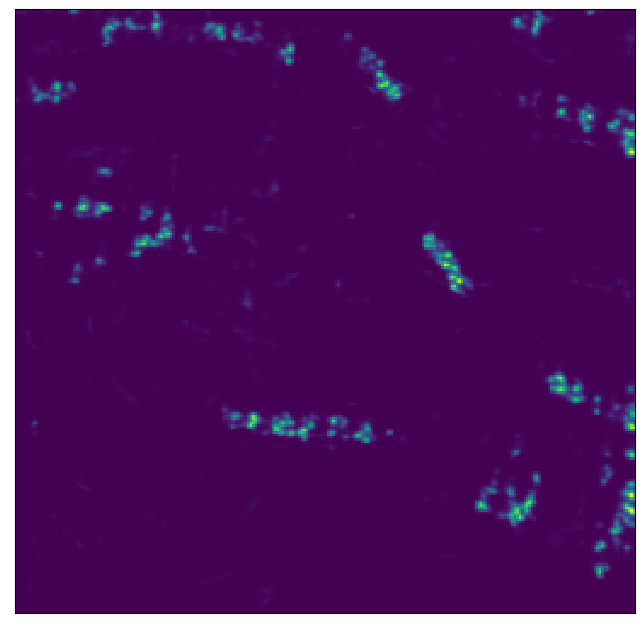} &
  \includegraphics[width=.23\textwidth]{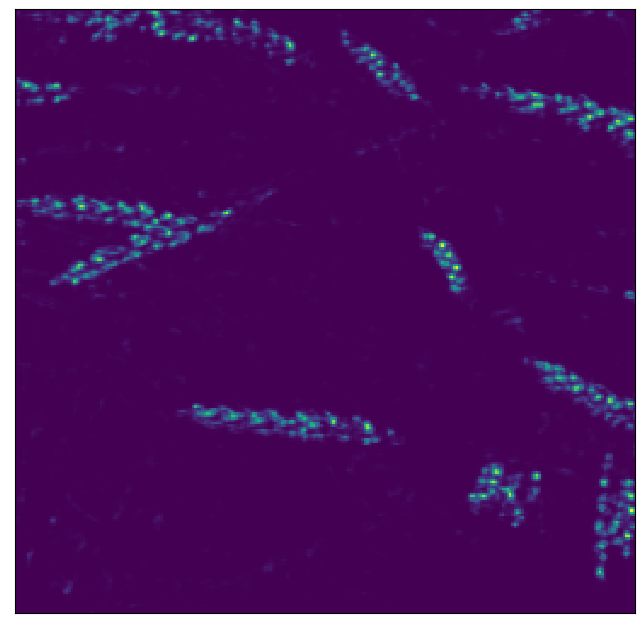} \\
  \includegraphics[width=.23\textwidth]{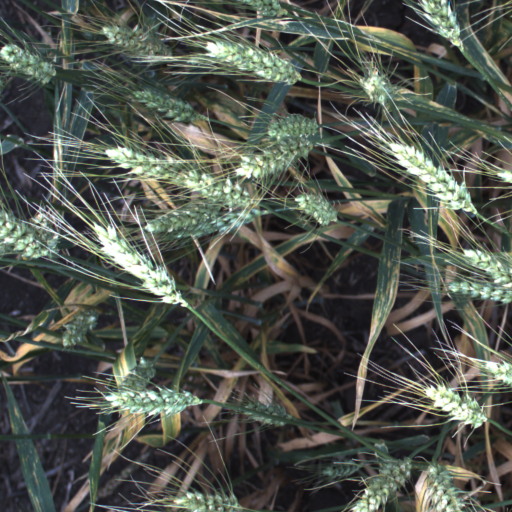} &
  \includegraphics[width=.23\textwidth]{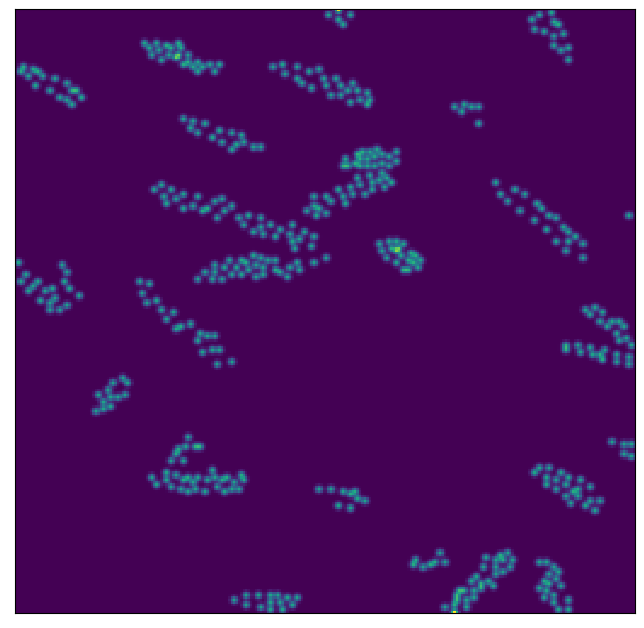} &
  \includegraphics[width=.23\textwidth]{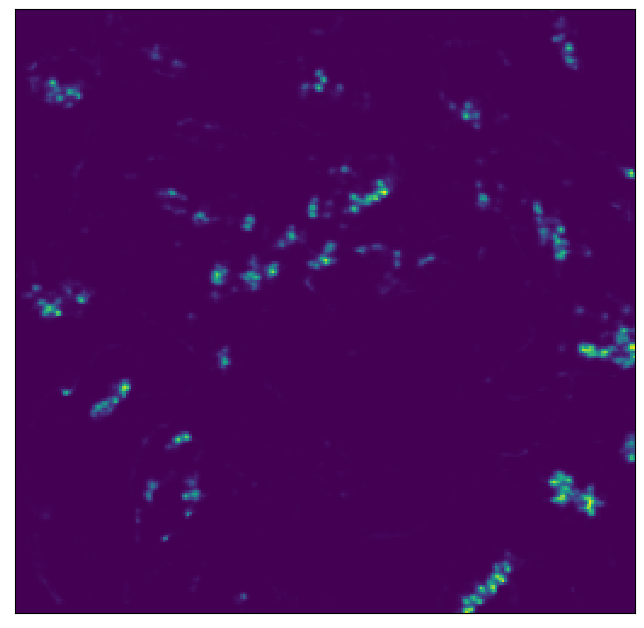} &
  \includegraphics[width=.23\textwidth]{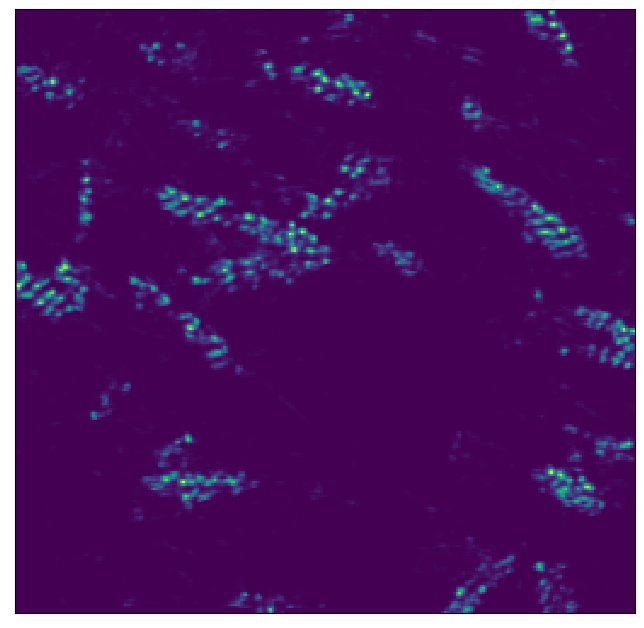} \\
  \includegraphics[width=.23\textwidth]{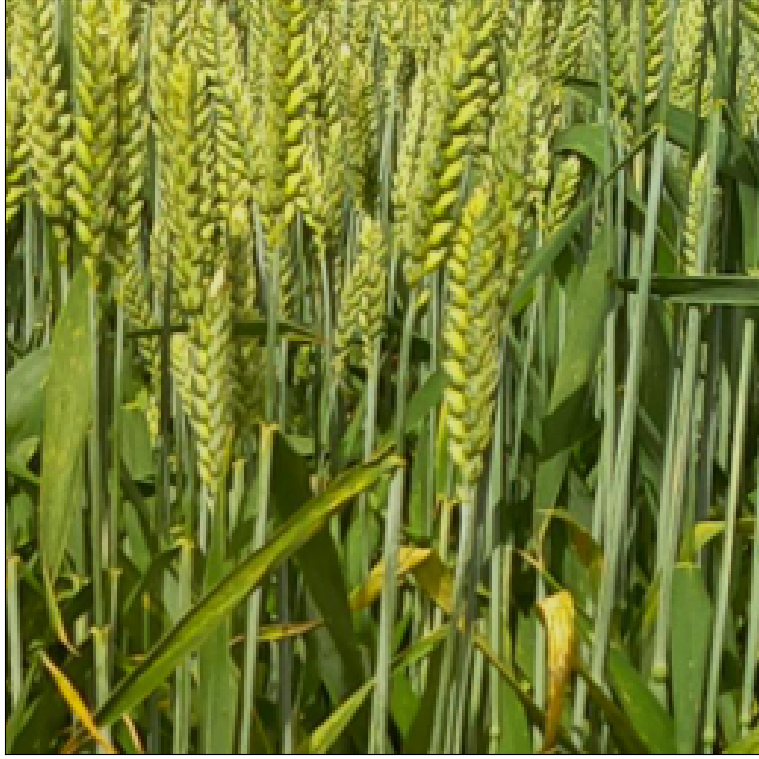} &
  \includegraphics[width=.23\textwidth]{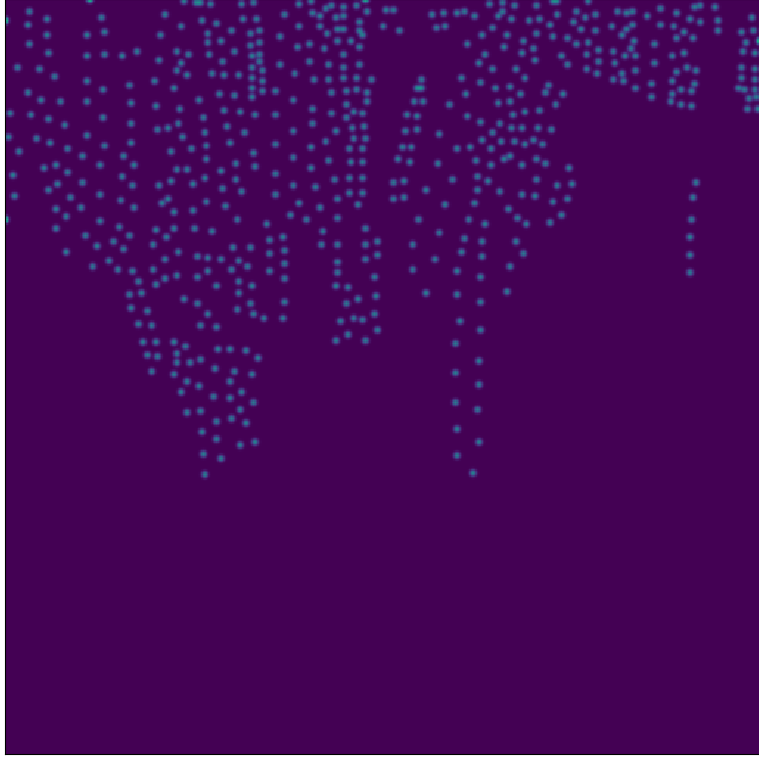} &
  \includegraphics[width=.23\textwidth]{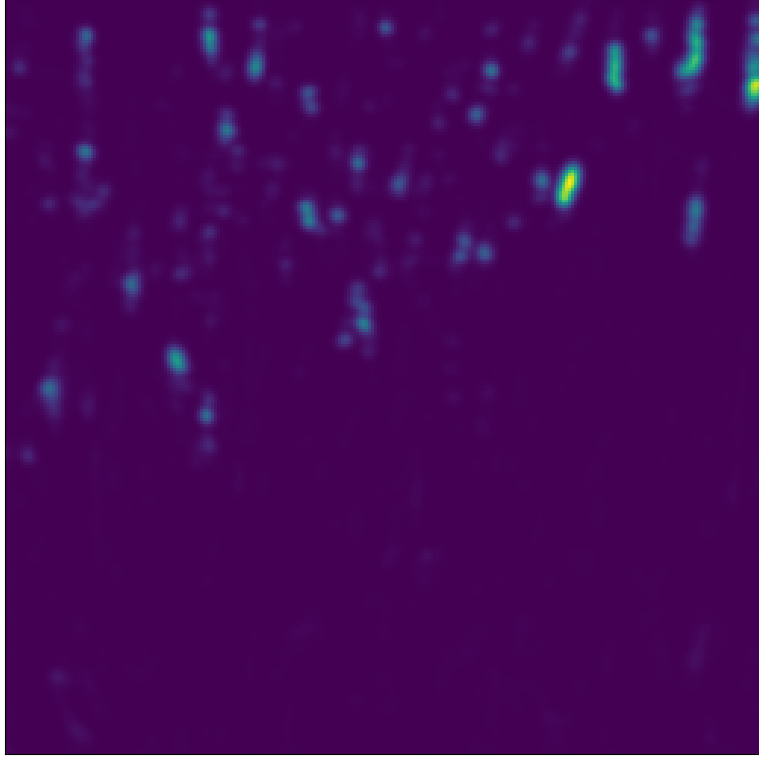} &
  \includegraphics[width=.23\textwidth]{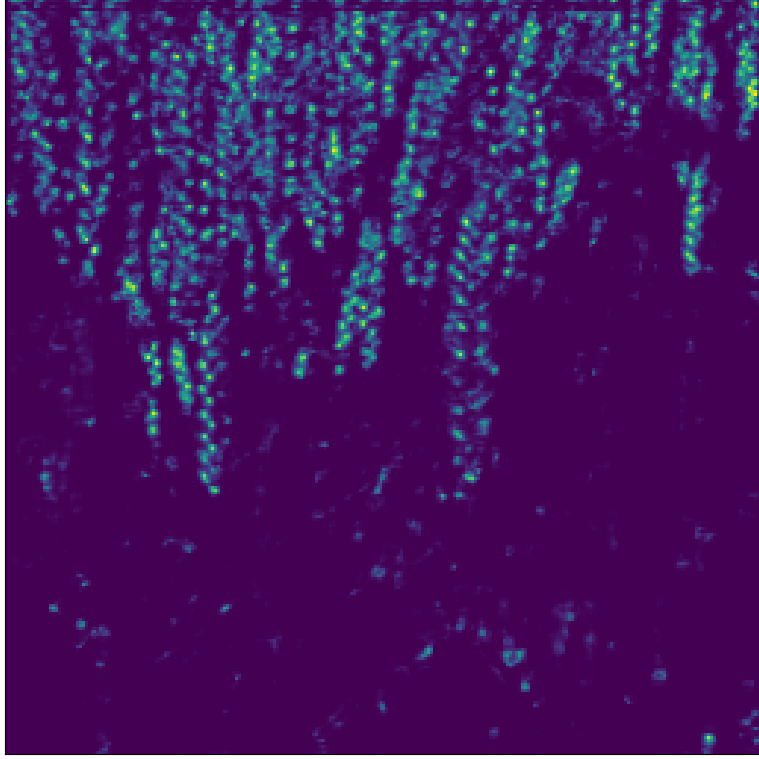} \\
  \includegraphics[width=.23\textwidth]{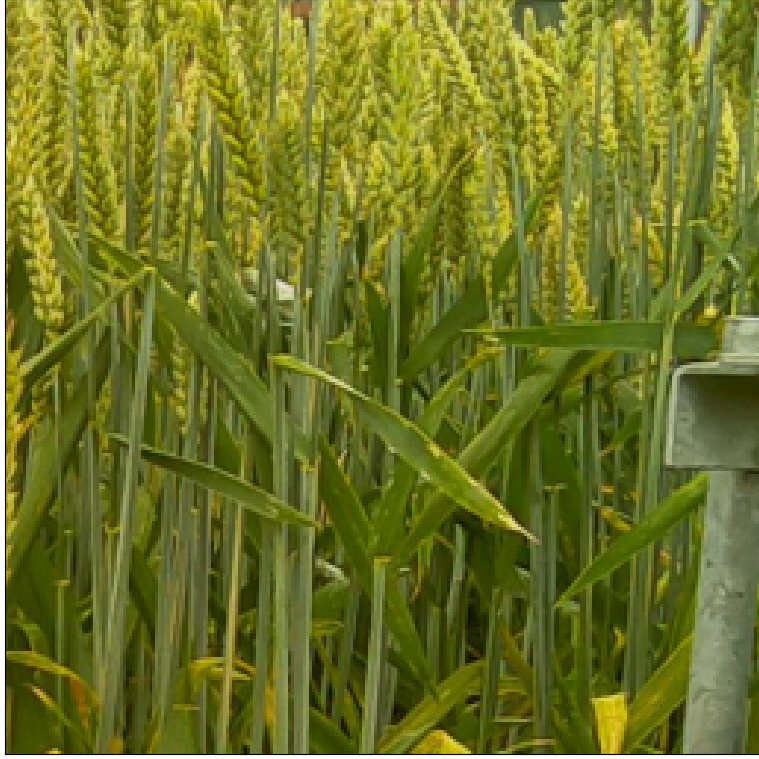} &
  \includegraphics[width=.23\textwidth]{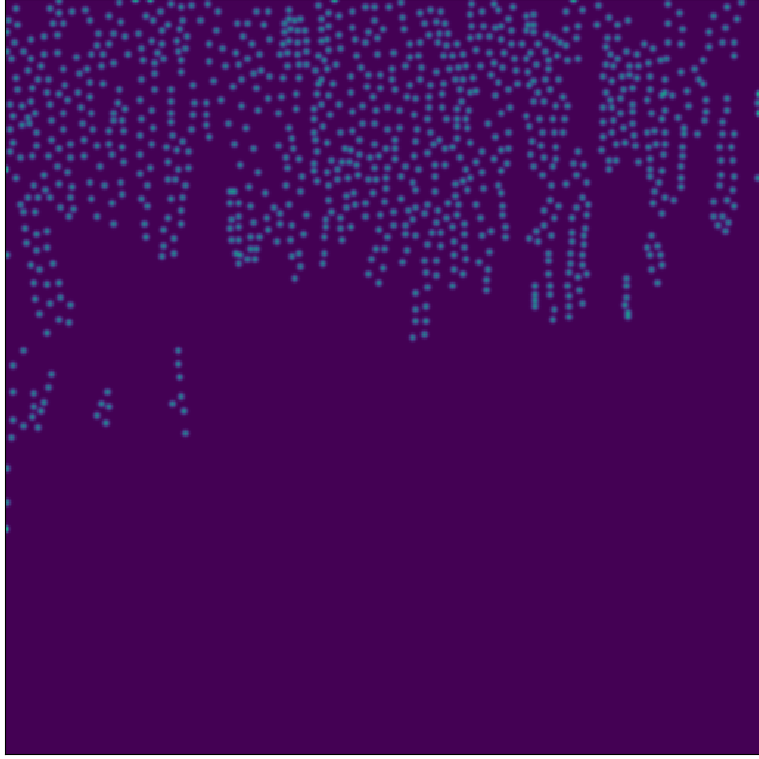} &
  \includegraphics[width=.23\textwidth]{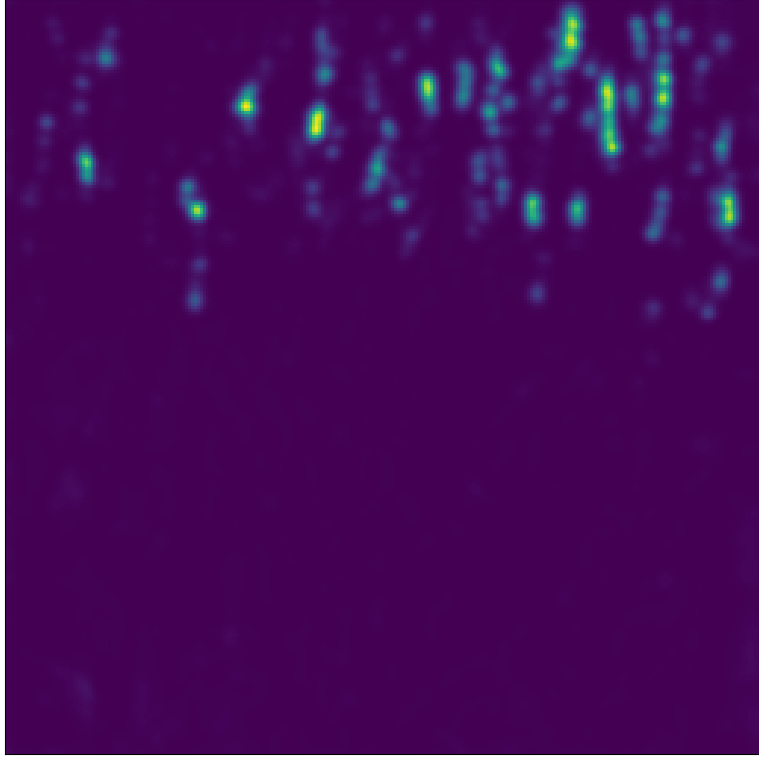} &
  \includegraphics[width=.23\textwidth]{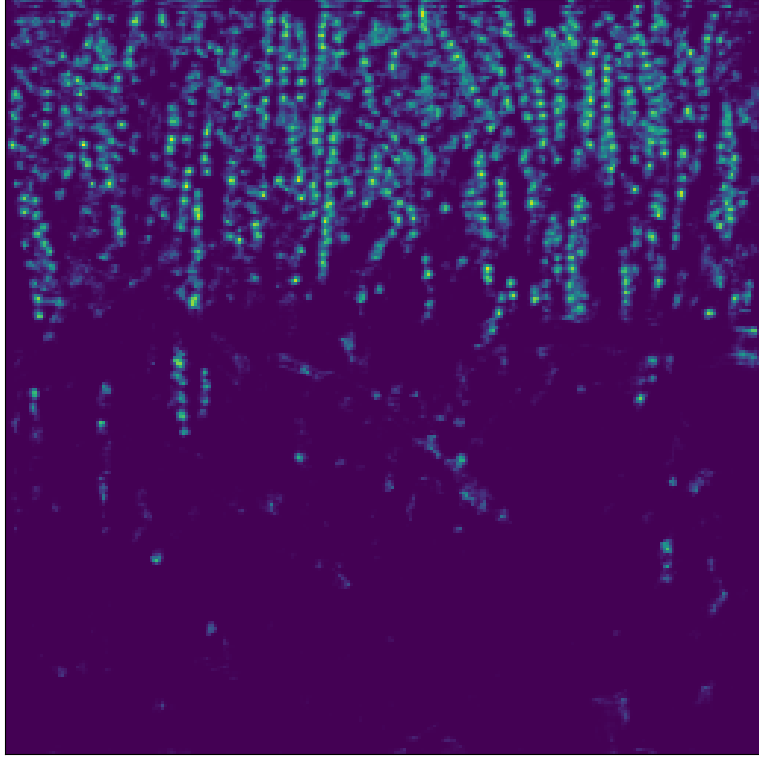} 
\end{tabular}%
\end{center}
\caption{Qualitative results for density map estimation. The first two rows are sampled from the Global Wheat Head Dataset and the last two are $512\times512$ patches extracted from CropQuant dataset (a) Input images. (b) Ground truth. (c) Predicted density map from the baseline model. (d) Predicted density map from the proposed model.}
\label{fig:spikelet_results}  
\end{figure}

Table~\ref{table:wheat_counting_results} summarizes the results from the wheat spikelet counting adaptation experiment. Figure~\ref{fig:spikelet_results} provides a qualitative comparison of example density maps output by the baseline model and by the proposed model.

\noindent \textbf{ACID to GWHD}: On the GWHD, our method reduced the MAE by $67.8\%$ and the RMSE by $68.7\%$ as compared to the baseline model without adaptation. The proposed method also achieved an $R^2$ value of 0.66 on the target domain. 

\noindent \textbf{ACID to CropQuant}: 
On patches extracted from the \textit{CropQuant} dataset, our method reduced the MAE by $59.3\%$ and the RMSE by $58.0\%$ as compared to the baseline model. The proposed method was also compared to two supervised methods from  \cite{alkhudaydi2019spikeletfcn}. One was trained solely on the target domain (from scratch), and the other was pre-trained on the ACID dataset and fine-tuned on the target domain (fine-tuned).
Our model outperformed the SpikeletFCN trained from scratch by $63.72\%$, while the fine-tuned SpikeletFCN model provides better performance than the proposed method. However, both of the reference methods require annotated datasets in the target domain while ours does not. 

\subsection{Leaf Counting}

\begin{figure}
\begin{center}
\begin{tabular}{cccc}
(a) Input & (b) Baseline & (c) Ours \\
  \includegraphics[width=.33\textwidth]{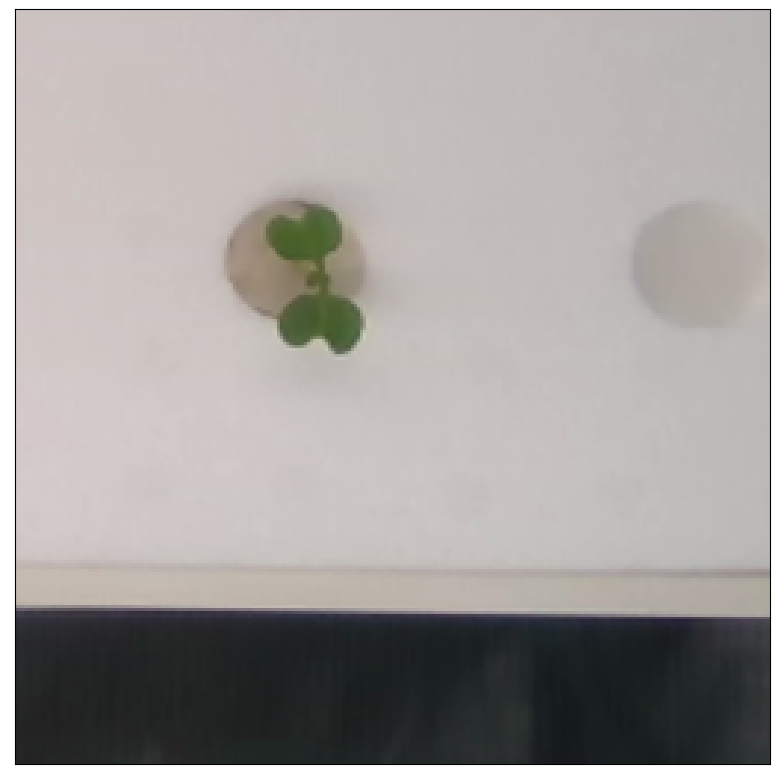} &
  \includegraphics[width=.33\textwidth]{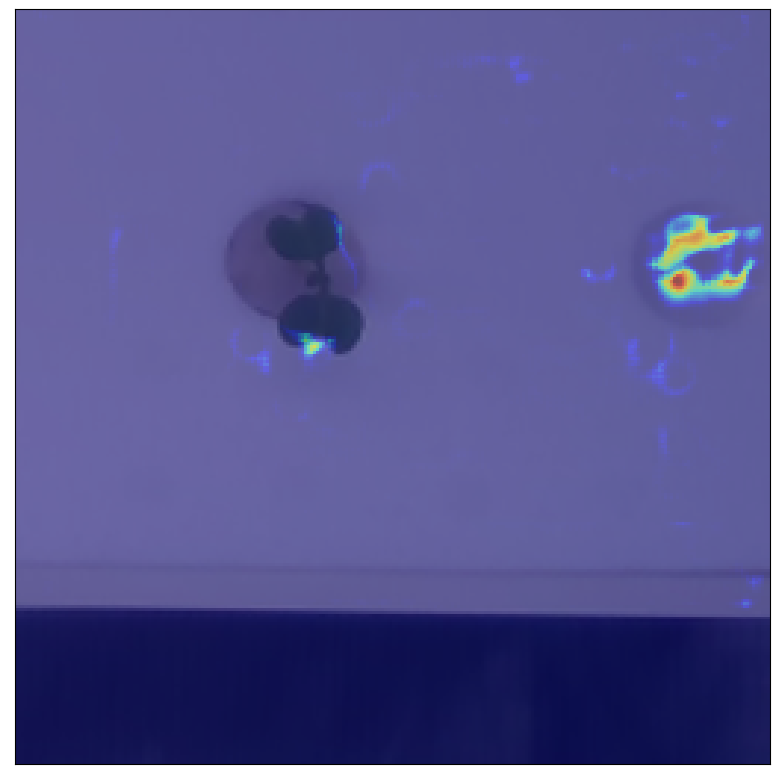} &
  \includegraphics[width=.33\textwidth]{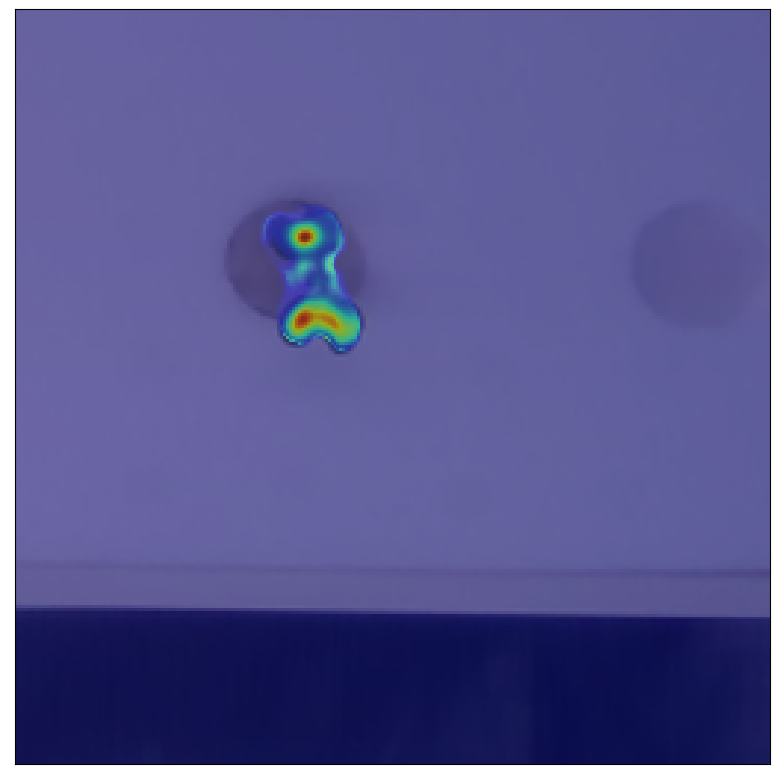} \\
  \includegraphics[width=.33\textwidth]{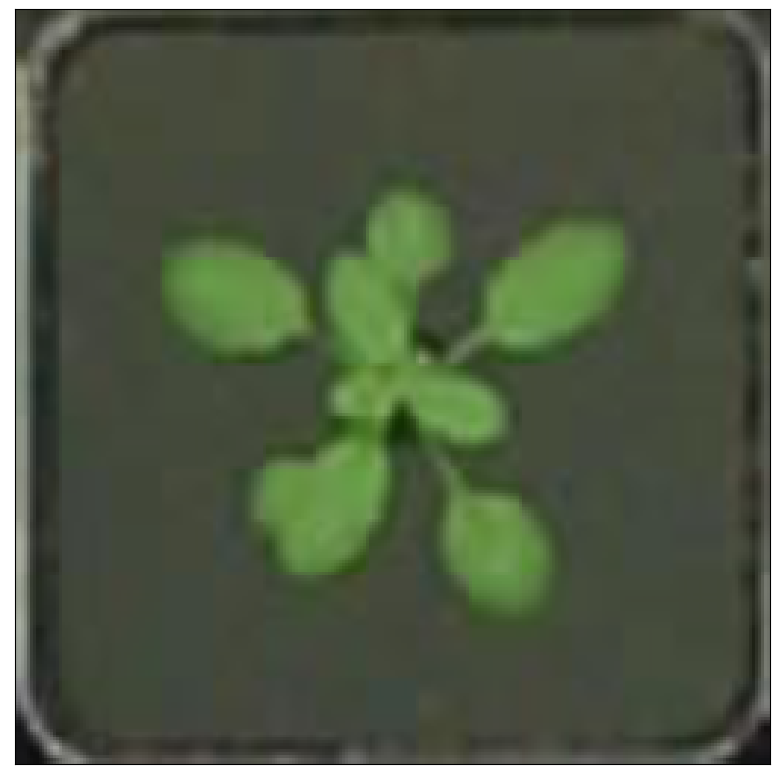} &
  \includegraphics[width=.33\textwidth]{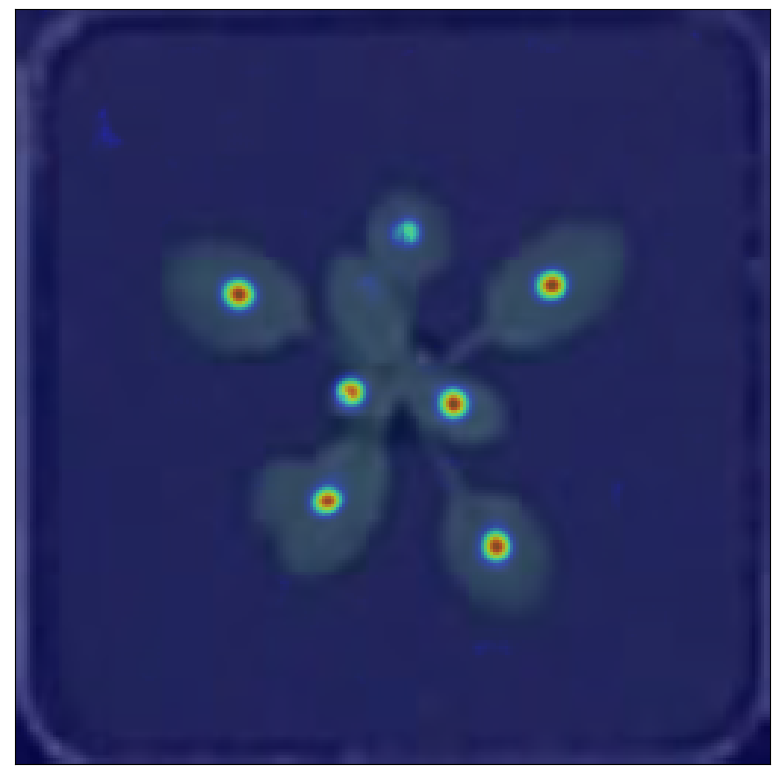} &
  \includegraphics[width=.33\textwidth]{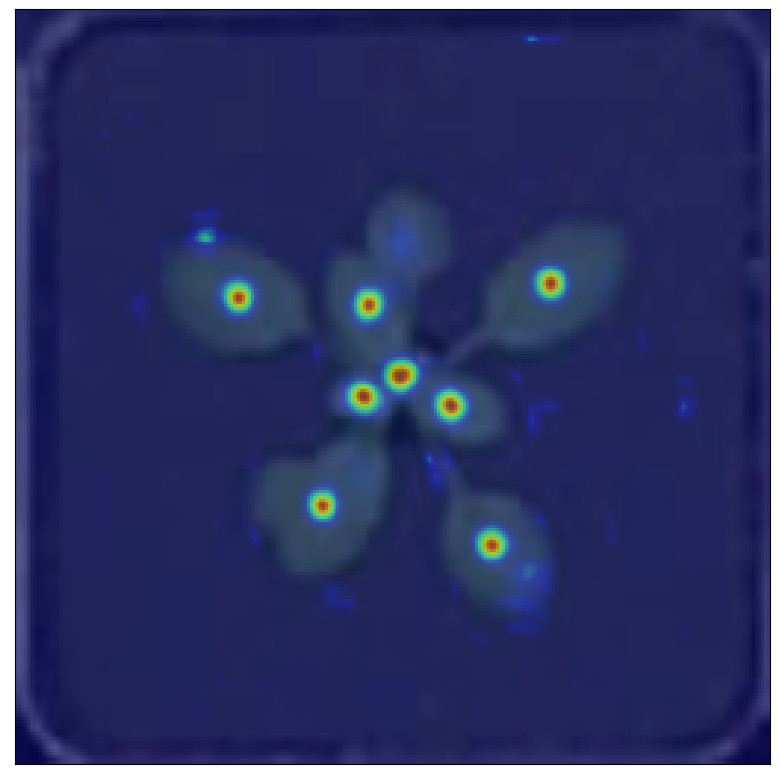} 
\end{tabular}%
\end{center}
\caption{
Sample input images (left) and density map estimations generated from the baseline model (middle) and from the adapted model (right) in the leaf counting task. Top row: \textbf{CVPPP to KOMATSUNA} experiment. Bottom row: \textbf{CVPPP to MSU-PID} experiment.}
\label{fig:komatsuna_test}
\end{figure}

Figure~\ref{fig:komatsuna_test} provides sample images taken from the KOMOTSUNA and MSU-PID datasets and shows that the proposed method accurately produces hot spots in the density map at the center of each leaf. 

\setlength{\tabcolsep}{4pt}
\begin{table}[h]
\begin{center}
\caption{Domain adaptation results for the leaf counting tasks. XE is cross-entropy loss, LS is least-squares loss,
$\downarrow$ denotes lower is better, $\uparrow$ denotes higher is better.}
\label{table:leaf_counting_results}
\begin{tabular}{ccccc}
\hline\noalign{\smallskip}
 &$DiC \downarrow$ & $|DiC| \downarrow$ & $\% \uparrow$ & $MSE \downarrow$\\
\noalign{\smallskip}
\hline
\noalign{\smallskip}
\textbf{CVPPP to KOMATSUNA} \\
\noalign{\smallskip}
Baseline (No Adaptation) & 4.09 (1.32) & 4.09 (1.32) & 0 & 18.49 \\
Giuffrida, et al. \cite{valerio2019leaf} (XE) &  -0.78 (1.12) & 1.04 (0.87) & 26 & 1.84\\
Giuffrida, et al. \cite{valerio2019leaf} (LS) &  -3.72 (1.93) & 3.72 (1.93) & 2 & 17.5\\
Our method & -0.95 (2.09) & 1.56 (1.67) & 29.33 & 5.26 \\
\noalign{\smallskip}
\hline
\noalign{\smallskip}
\textbf{CVPPP to MSU-PID} \\
\noalign{\smallskip}
Baseline (No Adaptation) & 1.21 (2.04) & 1.83 (1.52) & 0 & 5.65\\
Giuffrida, et al. \cite{valerio2019leaf} (XE) &  -0.81 (2.03) & 1.68 (1.39) & 20 & 4.78\\
Giuffrida, et al. \cite{valerio2019leaf} (LS) &  -0.39 (1.49) & 1.18 (0.98) & 26 & 2.36\\
Our method & 1.17 (1.85) & 1.63 (1.62) & 23.96 & 4.25 \\
\hline
\end{tabular}
\end{center}
\end{table}
\setlength{\tabcolsep}{1.4pt}

\noindent \textbf{CVPPP to KOMATSUNA}: 
We report results from the baseline model trained without adaptation, a previously proposed domain adaptation method \cite{valerio2019leaf}, and the proposed method (Table~\ref{table:leaf_counting_results}). The adapted model resulted in a $71.6\%$ drop in MSE, while the percentage agreement increased from $0\%$ to $29.33\%$. Additionally, we compared the proposed approach with a previous domain adaptation technique proposed by \cite{valerio2019leaf}. Our method outperformed the previous result when using the least-squares adversarial loss (LS), and achieves similar performance when using the cross-entropy loss (XE).


\begin{figure}
\begin{center}
\begin{tabular}{cc}
  \includegraphics[width=.5\textwidth]{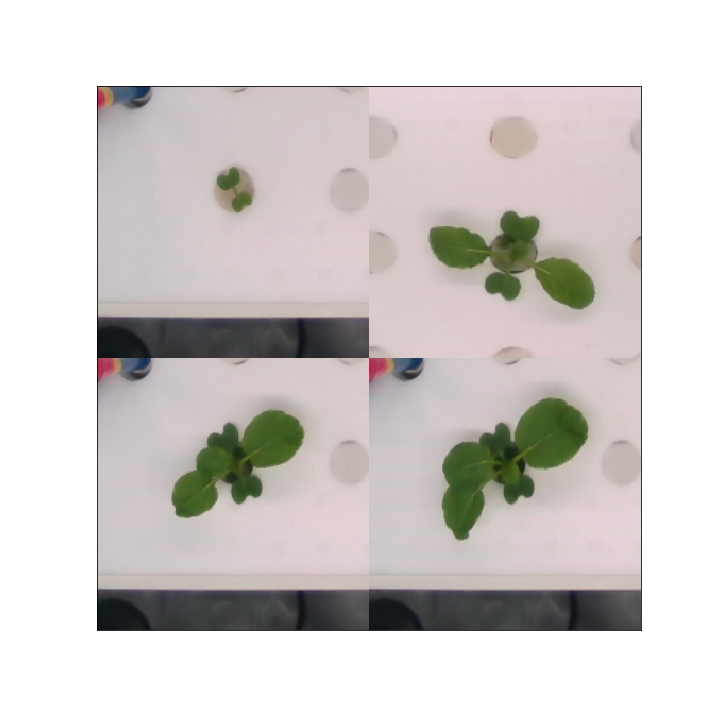} &
  \includegraphics[width=.5\textwidth]{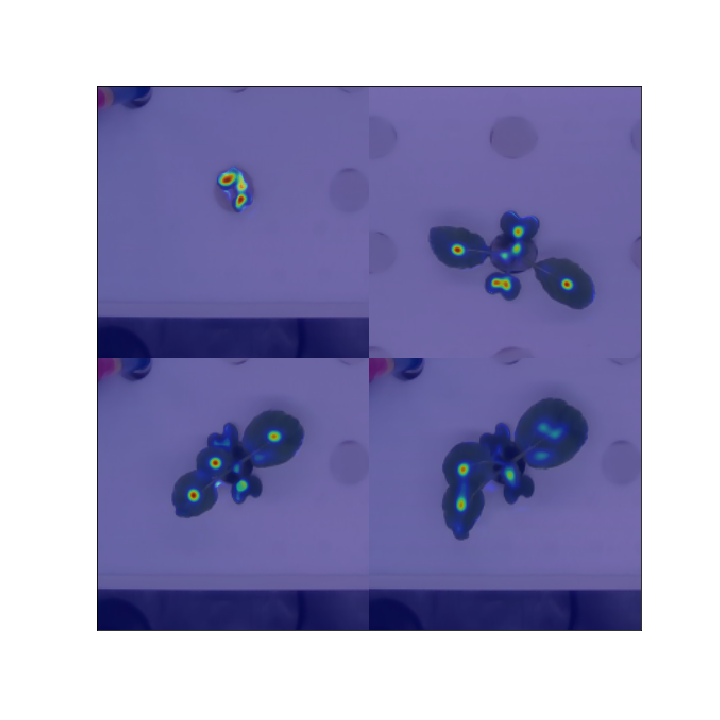} 
\end{tabular}%
\end{center}
\caption{
Sample composite image with multiple plants (left) and density map estimation (right) generated from the model trained to adapt leaf counting using domain and target datasets with one plant per image.
}
\label{fig:komatsuna_multiple}
\end{figure}

\noindent \textbf{CVPPP to MSU-PID}: 
On the MSU-PID dataset, our method outperformed the no-adaptation baseline model in all metrics (Table~\ref{table:leaf_counting_results}). The proposed model provided a $23.9\%$ increase in percentage agreement and a $10.93\%$ decrease in $|DiC|$. The proposed method outperformed the previous domain adaptation technique in this task when using the least-squares (LS) loss function, but under-performed it when using the cross-entropy (XE) loss. 

\noindent \textbf{Composite-KOMATSUNA}: Applying the model from \textit{CVPPP to KOMATSUNA} to composite images of multiple plants without retraining resulted in a mean $DiC$ of $9.89$ and mean $|DiC|$ of $9.90$. Figure~\ref{fig:komatsuna_multiple} shows a sample output from our model on the composite dataset. These results show that our model does not rely on assumptions about the distribution of object counts in the source domain, whereas previous work~\cite{valerio2019leaf} used a KL divergence loss between the distribution of leaf counts in the source domain and the predictions of their model to avoid posterior collapse.

\section{Summary}
In this paper, we attempted to address performance problems that arise in image-based plant phenotyping due to domain shift. Our study focused specifically on density estimation based object counting, which is a common method for plant organ counting. We proposed a custom Domain-Adversarial Neural Network architecture and trained it using the unsupervised domain adaptation technique for density estimation based organ counting. Our evaluation showed performance improvements compared to baseline models trained without domain adaptation for both wheat spikelet and leaf counting. For leaf counting, our results show similar performance to a previously proposed domain adaptation approach without the need to manually select a loss function for each dataset. This study is also the first to investigate the use of domain adaptation from an indoor source dataset to an outdoor target dataset. This may be a viable method for many plant phenotyping contexts, such as plant breeding experiments with both controlled environment and field trials. We plan to confirm this indoor-to-outdoor adaptation approach through additional testing with different plant species, different organ counting tasks, and additional field conditions, as future work.

\section{Acknowledgement}

This research was undertaken thanks in part to funding from the Canada First Research Excellence Fund. 

%
%
\bibliographystyle{splncs04}
\bibliography{paper}
\end{document}